\RequirePackage{fix-cm}
\documentclass[smallcondensed]{svjour3}     
\smartqed  
\usepackage{graphicx}
\usepackage{natbib}

\usepackage{amsmath}
\usepackage{enumitem}
\usepackage{float}
\usepackage{hyperref}
\usepackage{amsmath,amssymb,amsfonts}
\usepackage{algorithm, algpseudocode}
\algnewcommand\algorithmicforeach{\textbf{for each}}
\algdef{S}[FOR]{ForEach}[1]{\algorithmicforeach\ #1\ \algorithmicdo}
\usepackage{color}

\usepackage{makecell}
\begin{document}

\title{Decentralised construction of a global coordinate system in a large swarm of minimalistic robots}
\titlerunning{Distributed self-localisation in a large swarm of minimalistic robots}        

\author{Michal Pluhacek \and Simon Garnier \and Andreagiovanni Reina}


\institute{M. Pluhacek \at
            Faculty of Applied Informatics,
            Tomas Bata University in Zlín,
            Zlín, Czech Republic\\
              \email{pluhacek@utb.cz}           
        \and
        S. Garnier \at
        New Jersey Institute of Technology, NJ, USA\\
        \email{garnier@njit.edu}
        \and
        A. Reina \at
        IRIDIA, Universit\'{e} Libre de Bruxelles, Brussels, Belgium\\
        Sheffield Robotics, University of Sheffield, Sheffield, UK\\
        \email{andreagiovanni.reina@ulb.be}
}


\maketitle

\begin{abstract}
Collective intelligence and autonomy of robot swarms can be improved by enabling the individual robots to become aware they are the constituent units of a larger whole and what is their role. In this study, we present an algorithm to enable positional self-awareness in a swarm of minimalistic error-prone robots which can only locally broadcast messages and estimate the distance from their neighbours. Despite being unable to measure the bearing of incoming messages, the robots running our algorithm can calculate their position within a swarm deployed in a regular formation. We show through experiments with up to 200 Kilobot robots that such positional self-awareness can be employed by the robots to create a shared coordinate system and dynamically self-assign location-dependent tasks. Our solution has fewer requirements than state-of-the-art algorithms and contains collective noise-filtering mechanisms. Therefore, it has an extended range of robotic platforms on which it can run. All robots are interchangeable, run the same code, and do not need any prior knowledge. Through our algorithm, robots reach collective synchronisation, and can autonomously become self-aware of the swarm's spatial configuration and their position within it.

\end{abstract}

\section{Introduction}
\label{intro}

Coordination in natural and artificial collective systems is, fundamentally, a spatiotemporal problem. For two or more agents to coordinate their work, each must have a sense, even if imperfect, of when and where others' actions have taken place relative to its individual frame of reference. For instance, flocking in birds or schooling in fish is only possible if each individual can quickly adjust their movement (a spatiotemporal attribute) to the movements of their immediate neighbours (another spatiotemporal attribute). Even in fixed groups of interacting agents, for instance, a network of sensors and actuators in a building, the topology (spatial) and timing (temporal) of the interactions will determine the type (e.g., synchronisation, oscillation) and the quality of the collective coordination.

In the context of swarm robotics \citep{hamann_swarm_2018}, a field of research studying the coordination of large groups of simple autonomous robots, the temporal aspect of coordination is relatively easy to solve. Indeed, the microprocessors that control the behaviour of the robots are, essentially, clocks. They provide each robot with an internal temporal frame of reference against which it can log the events that it perceives from its environment. The spatial aspect, that is where an event happens relative to a focal individual, is, however, less straightforward. Traditionally, it is solved using either an external frame of reference (e.g. a Global Positioning System) or implementing a "mental" mapping mechanism within the robot's controller \citep{pritsker_introduction_1984}. The former is usually very precise, but GPS signals are not always accessible to the robotic agents (e.g. when they are blocked by obstacles) and their use runs somewhat contrary to the full-autonomy goal of swarm robotics. The latter provides increased autonomy to the robotic swarm but requires significant processing power and memory storage, which may not be available on small or microscopic robotic agents.

Here, we propose an alternative approach using a fully decentralised mechanism that can be implemented in autonomous robots with limited capabilities. In particular, our approach allows a group of robots to build a global coordinate system without requiring external reference information, preset origin, or predetermined roles for the robotic agents. It is designed to work on any robotic platform, even with minimal, undirected communication abilities and noisy distance sensors. To demonstrate the feasibility of our approach, we implemented it using the well-known, minimalist Kilobot platform. We also show that it scales up to large robotic swarms by performing validation experiments with up to 200 real and 1,000 simulated Kilobot units.

The rest of the manuscript will be organised as follows. First, we will give a general description of our approach and compare it with existing approaches in the literature, with a focus on minimalist robots (Section \ref{sec:stateart}). In particular, we will highlight the strengths and limitations of our approach with respect to the existing ones. Then, we will provide a detailed description of the proposed algorithm and its implementation in the Kilobot platform (Section \ref{sec:method}). This will be followed by a description of the results of multiple experiments with real and simulated Kilobot swarms demonstrating the capabilities and scalability of the approach (Section \ref{sec:experiments}), before we offer our conclusions on the possibilities that it offers and on directions for future studies (Section \ref{sec:conclusion}).

\section{Previous work}
\label{sec:stateart}

\begin{figure}[t]
\begin{tabular}{cc}
  \includegraphics[width=0.45\textwidth]{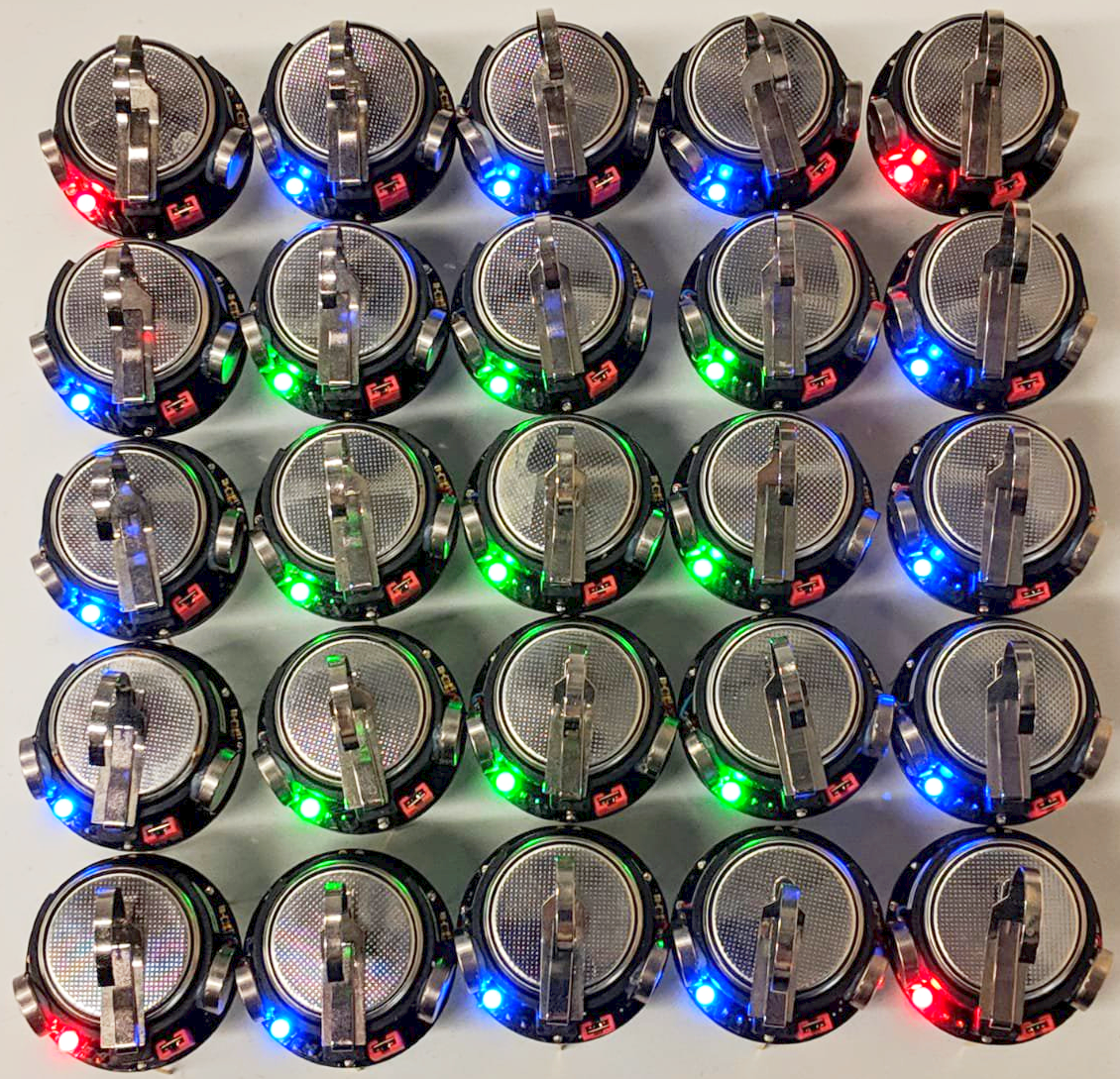}
 & 
  \includegraphics[width=0.45\textwidth]{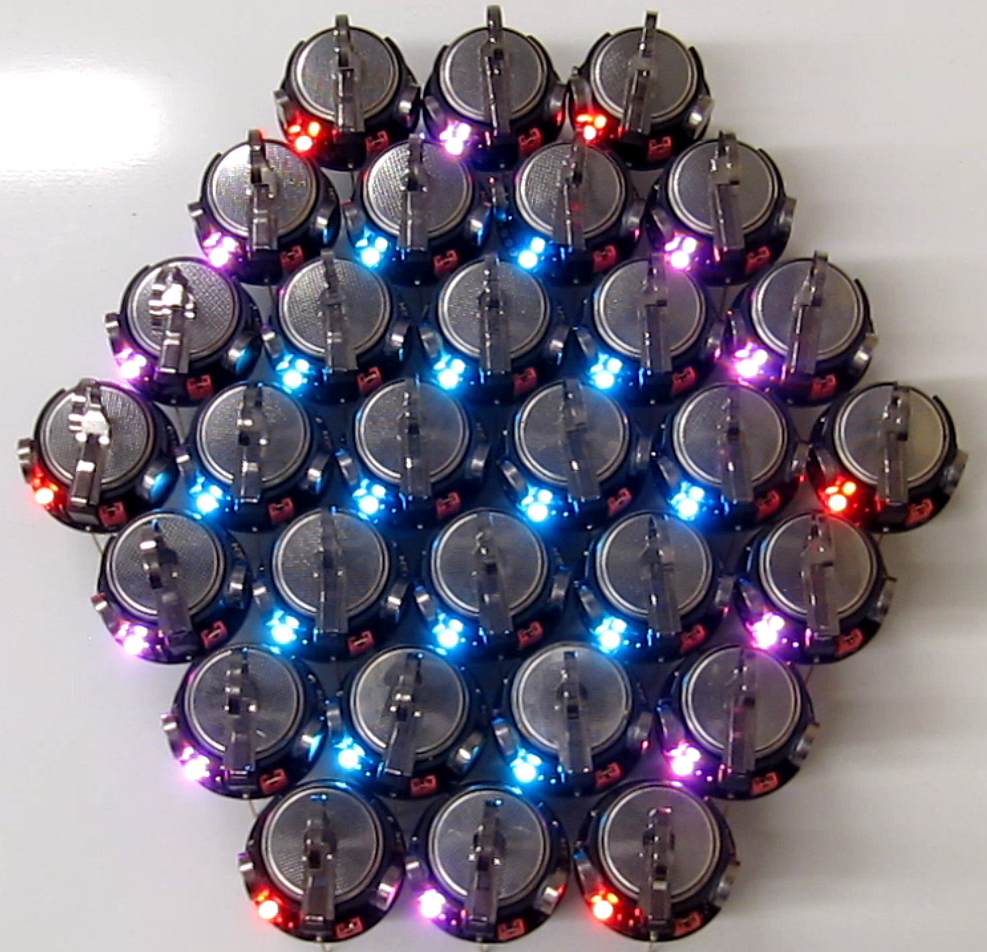}

\\
a) $5\times5$ rectangular lattice & b) $4\times3\times4$ hexagonal lattice \\
\end{tabular}
\caption{Two Kilobot swarms organised in two regular formations with two different topologies. In both cases, the robots by running the Routine R1 (Sec.\,\ref{sec:routineR1}) are able to identify their neighbourhood on the two distinct regular lattices and compute their position group, i.e. CORNER position displayed with a red light, BORDER position with a blue light in (a) and a magenta light in (b), and MIDDLE position with a green light in (a) and a cyan light in (b).}
    \label{fig:neighborhood_examples}
\end{figure}

Through our algorithm, robots can autonomously self-localise within a group and dynamically self-assign roles depending on their position. The algorithm works under the assumption that the robots are organised in a regular formation, in either a rectangular or hexagonal lattice, as illustrated in two examples in Figure \ref{fig:neighborhood_examples}. Having the robots organised in such regular formations can be particularly convenient for logistics reasons. Indeed, robots are normally stored, charged, transported, and deployed in regular formations: squared and rectangular lattices simplify the working logistics and hexagonal lattices maximise the packing density of robots with circular bodies. 
In addition to simplifying the transport and deployment logistics, having robots in regular formations can be a requirement for the successful execution of the collective task, for instance, the coherent motion of multi-robot aggregates \citep{pratissoli_soft-bodied_2019} and light pattern display \citep{alhafnawi_robotic_2020}. In these works, spatiotemporal coordination was achieved by manually providing the robots with information about their relative position within the formation. Our algorithm, by enabling the robots to self-localise, increases the system's autonomy.

There are other studies that proposed decentralised algorithms for the construction of coordinate systems and self-localisation in robot swarms. Our solution is able to work with fewer requirements and on simpler robots than state-of-the-art methods. In particular, there are a few decentralised algorithms \citep{beal_organizing_2013,sahin_swarm-bot_2002,guo_swarm_2011,coppola_provable_2019,mathews_mergeable_2017,wang_decentralized_2021,Batra2022,Li2018,Klingner2019} that allow each robot to compute its relative positioning with respect to the rest of the swarm through the use of distance and directional information about neighbours---i.e.~each robot is able to know the relative location of other robots nearby---and in some cases of a global reference orientation (e.g., a compass). On the one hand, our algorithm can work on simpler robots only equipped with noisy distance sensors and transceivers for local broadcasting of small messages. Therefore, with our solution, neighbours' bearing is not needed, making possible its implementation on Kilobots and other minimalist robotic platforms. On the other hand, our solution requires the deployment of the robots in a regular formation while the robots could operate in arbitrary arrangements in the studies cited above, as long as the robots could communicate with each other. In addition, previous studies that implemented coordinate system construction on Kilobot swarms also required the robot to be deployed in a regular formation, in their case in a hexagonal lattice \citep{rubenstein_programmable_2014, gauci_programmable_2018}. However, in their case, a subset of the robots ran a different code than the rest of the swarm and needed a precise initial placement to form the coordinate system origin and axes orientation. In our swarm, all robots run identical code and do not need any pre-configuration. Every robot can be replaced with any other and their relative position interchanged without compromising the collective behaviour. Additionally, our algorithm executes simpler mathematical operations than the previous methods by \cite{rubenstein_programmable_2014} and \cite{gauci_programmable_2018}, which can be better suited for minimal computing devices.

\section{Self-organised construction of a coordinate system}
\label{sec:method}

\begin{figure}[t]
\centering
  \includegraphics[width=0.85\textwidth]{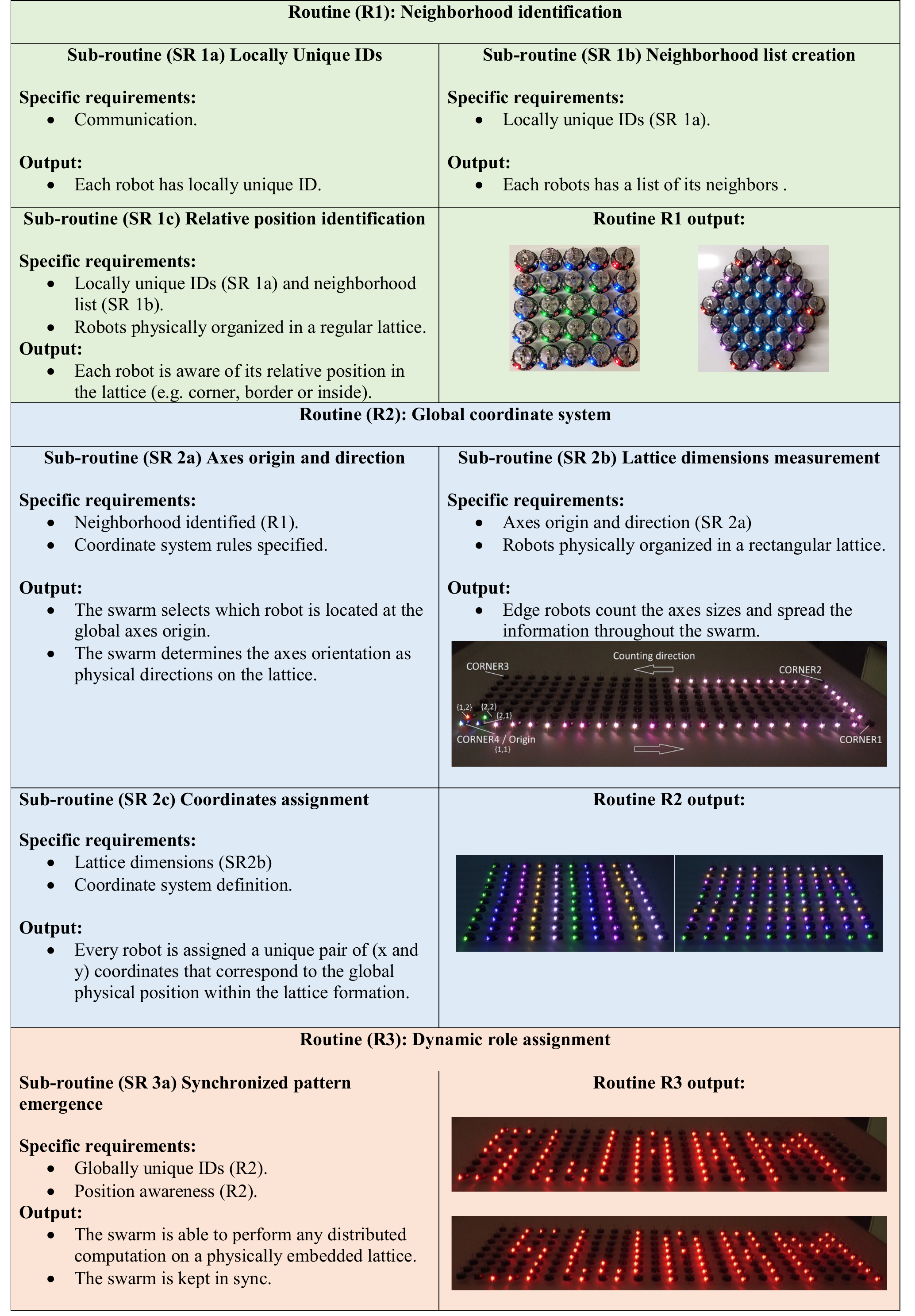}
\caption{Overview of the three routines to (R1) allow every robot to identify its lattice neighbourhood, (R2) construct a swarm-level coordinate system, and (R3) dynamically assign roles to robots in specific positions within the formation. All routines consist of decentralised algorithms to let the robot reach global coordination in a self-organised way. Each routine comprises one or more sub-routines (SR).}
\label{fig:overview}
\end{figure}

In order to build a shared coordinate system, the robots run a sequence of three \textit{Routines}:
\begin{enumerate}[label=R\arabic*.,leftmargin=1cm,itemsep=0.1cm]
\item Neighbourhood construction
\item Coordinate system construction
\item Synchronised dynamic role assignment
\end{enumerate}
Each routine is composed of sub-routines that we describe in detail in this section. Figure \ref{fig:overview} gives an overview of the full process. Our three routines are designed to be run by each robot comprising the swarm. Every robot runs identical code, therefore, there is no requirement to pre-assign roles before deployment and any robot can replace any other. Despite the robots relying solely on local and error-prone communication, the execution of routines R1 and R2 allows the swarm to build a global-level coordinate system and enables every robot to self-localise within the shared coordinate system. Such capabilities can, then, allow the robots to self-assign roles based on their position within the swarm and on a shared time clock, as showcased with routine R3.
The routines' algorithms are presented in a generic form and are designed to work on both rectangular and hexagonal regular lattices, with exception of routine R2 which has been tailored to build a coordinate system on rectangular lattices only. In fact, the coordinate system in rectangular and hexagonal lattices is structurally different \citep{snyder_coordinate_1999}, therefore, routine R2 needs to be designed specifically for each type of lattice. For the purpose of this study, we show the construction of the coordinate system (R2) in rectangular lattices only.

Before describing the three routines, we explain how the robots can maintain a time clock synchronised among all the robots, which is a fundamental requirement for the successful execution of our routines.

\subsection{A synchronised swarm}
\label{seC:synch}
In our study, every robot is an autonomous entity that runs the same code and exchanges messages with close neighbours. In several parts of the process, the transition from one subroutine to the next one is based on a shared time clock. However, without any closed-loop control, the independent clocks of a large number of devices can desynchronise over time. On the contrary, implementing a communication protocol that maintains every device in step with the same global clock at all times can require either a global orchestrator or fast communication and large bandwidth. Therefore, in our system, we implement a hybrid solution that combines independent clocks and closed-loop synchronisation: the robots independently measure time during limited periods and periodically exchange synchronisation messages to reset their clocks and avoid large drift.

\paragraph{Individual clocks.}
In our implementation, Kilobots are equipped with a microcontroller that runs the robot's control algorithm at a rate of approximately 32 Hz. The approximate time elapsed from an event (e.g., the start of the process) can thus be measured using the number of control loops that the robot executed (which in the Kilobot's firmware are expressed as \texttt{kilo\_ticks}). This approach is easy to implement and does not require communication among the robots, however, despite calibration by the producer, the Kilobots' clocks can quickly desynchronise over time. After a few minutes, the difference can become noticeable. Therefore, in our implementation, robots use their internal clock to estimate the time elapsed from the beginning of the current phase of the process (e.g., a subroutine) and synchronise their internal counter at every phase change.

\paragraph{Message-based synchronisation.}
When a robot reaches the given time limit according to its internal clock, it moves to the next phase of the algorithm. As soon as this happens, the robot broadcasts a synchronisation message to notify that the next algorithm phase has begun. Any robot that receives this synchronisation signal moves to the next phase and relays the message. As every phase runs for a relatively short time period, the accumulated difference among robot clocks is typically small. Therefore, in large swarms, it often happens that multiple robots independently reach the time limit according to their internal clock before receiving others' synchronisation messages. Thus, in our system, synchronisation signals are independently generated from multiple sources, decreasing the delay that might occur in spreading a message from a single source throughout a large swarm. This approach has been tested with up to 200 real Kilobots and 1000 simulated robots and the swarm has not desynchronised over extended periods of time nor accumulated any notable delay.

\subsection{Routine R1 -- Neighbourhood construction}
\label{sec:routineR1}
The routine \textit{Neighbourhood construction} (R1) aims to enable each robot to construct a list of its closest neighbours which can be identified with locally-unique IDs. The routine R1 is composed of three subroutines: SR1a Locally-unique ID assignment, SR1b Neighbour list creation, and SR1c Relative position identification.


\subsubsection{Subroutine SR1a -- Locally-unique IDs assignment}
A locally-unique ID is essential for neighbourhood construction as it allows robots to count the number of robots in their proximity and establish one-to-one communication with each neighbour. This subroutine is composed of two phases presented as pseudocode in Algorithm \ref{code:SR1b}. During the first phase, each robot generates a random ID that it draws from a given range. In our case, the Kilobots use an 8-bit integer in the range [0,255]. The robots broadcast their randomly selected ID. Each received ID is added to a blacklist of already used IDs. If a robot receives a message with an ID equal to its own ID, it selects a new random ID from the same range, excluding the IDs in the blacklist. After sufficient time has elapsed, in our case the time necessary to send about 20 messages (10\,s), see line \ref{line:timer1} of the Algorithm \ref{code:SR1b}, the subroutine moves to the second phase. In this phase, the robots aim to remove duplicated IDs among their neighbours who may not be in direct communication with each other. However, a robot with two neighbours that use the same ID cannot distinguish between them and this causes problems in subsequent subroutines. In this second phase, each robot generates a new random number and repeatedly broadcasts messages containing four pieces of information: its ID, the just-generated random number, and the ID and the random number received from one of its neighbours. Each new message contains information about a different neighbour, iteratively selected (lines \ref{line:new-message}-\ref{line:new-message2}). For each received message, the robot stores the received neighbour's ID and the additional random number in a list. The use of the additional random number is necessary to distinguish between its own ID included in the neighbour's message or the ID from another robot with the same value. The probability of random selection of both the same ID and the same random number reduces exponentially with the range size. For instance, for the Kilobots using two 1-byte numbers, the probability is smaller than 0.002\%. When a robot receives a message in which the third piece of information is equal to its ID and the fourth piece of information is not equal to its random number, it means that there are repeated IDs (line \ref{line:duplicated-id-check2}). Therefore, the robot adds its ID to the blacklist and self-assigns a new random ID. After sufficient time has elapsed, in our case the time to send about 30 messages (15\,s, line \ref{line:timer2}), the subroutine SR1a terminates and the subroutine SR1b begins.

\subsubsection{Subroutine SR1b -- Neighbour list creation}
In order to create a list of its neighbours, a robot must filter incoming messages based on the senders' distance. In our implementation, the Kilobots exchange messages through infrared (IR) messages and can estimate the sender distance using the IR signal strength \cite{rubenstein_kilobot_2012}. In agreement with the Kilobots' capabilities, we assume that the robots can only rely on distances without directional information, that is robots cannot know the relative angle of the sender, they can only know its distance.

Subroutine SR1b is composed of three phases, and its pseudocode is shown in Algorithm \ref{code:SR1b}. In the first phase, every robot continues, as it did during SR1a, to broadcast its locally-unique ID. Each robot measures the distance of the incoming messages and records the shortest distance $x$ (erroneous values smaller than the robot body-size are discarded, e.g. 33\,mm for the Kilobots). After a threshold time, 
the robots start the second phase. In our case, we combined the first phase of SR1b with the second phase of the subroutine SR1a (line \ref{line:min-distance} of Algorithm \ref{code:SR1b}), to speed up the process. Based on the shortest distance recorded, the robot computes the maximum neighbourhood radius $r$ which it uses to filter the incoming messages and create the neighbour list. A robot only adds to its neighbour list the ID of senders at a distance smaller than $r$.
Computing the radius $r$ as a function of the estimated shortest distance $x$ allows the algorithm to adapt to different spacing and grid layouts.

Subroutine SR1b enables the robots to create their neighbour list when they are organised in the regular lattice topologies of rectangular lattices (as in Fig. \ref{fig:neighborhood_examples}a) and equilateral triangular lattices, also known as hexagonal lattice (as in Fig. \ref{fig:neighborhood_examples}b). For the case of hexagonal structures, computing $r$ is easier as all robot neighbours are at approximately the same distance. Therefore, $r$ can be set to $r=(1+\epsilon)x$, where $\epsilon < 0.5$ is the proportion of tolerable error in placement/sensing. For the case of rectangular lattices we want to identify the Moore neighbourhood, therefore computing $r$ is more difficult as it must include robots at different distances: the neighbours on the diagonal are farther than the ones at the sides. Additionally, the rectangular lattices can have  different vertical and horizontal distances, that is the lattice's rows can be closer than the columns are, or vice versa. Given the robots' limitations (i.e. absence of directional information), subroutine SR1b can only work with an asymmetry between vertical and horizontal distances up to a given limit. The limit is given by the inequality $2x > \sqrt{x^2+y^2}$, where $x$ and $y$ are the inter-robot distances on the two dimensions (rows and columns). Assuming $x\le y$, the inequality states that twice the distance of one dimension should be larger than the distance on the diagonal of the rectangular grid. Thus, we have the constraint that $y<\sqrt{3}x$. If we also assume a placement/sensing error $\epsilon$, the inequality becomes $2x(1-\epsilon)> \sqrt{(x+\epsilon\,x)^2+(y+\epsilon\,y)^2}$, resulting to
\begin{equation}
    y<\frac{x \sqrt{3 \epsilon ^2-10 \epsilon +3}}{\sqrt{\epsilon ^2+2 \epsilon +1}}.
\end{equation}

\begin{figure}[h!]
\begin{tabular}{cccc}
\includegraphics[width=0.22\textwidth]{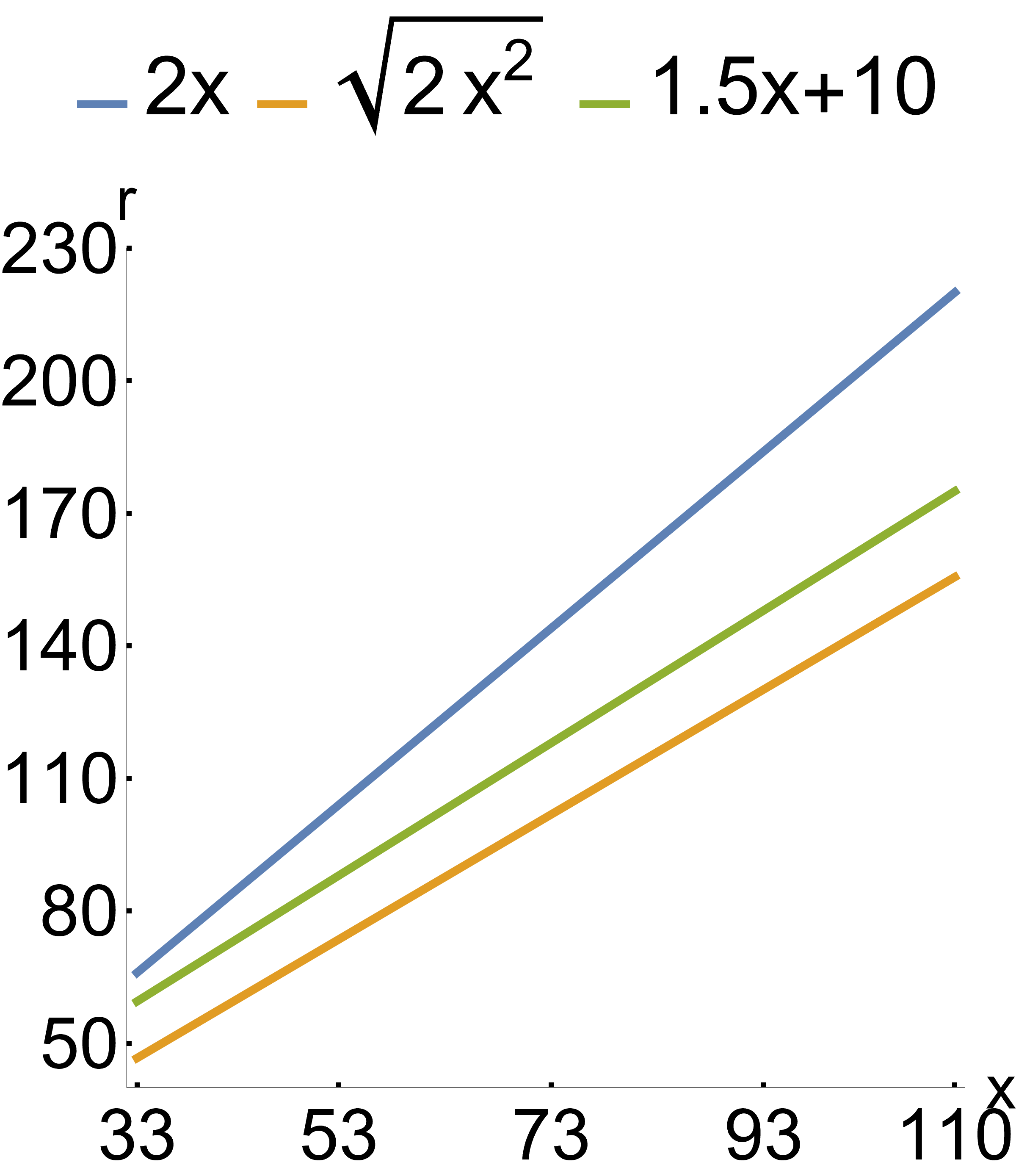}&
  \includegraphics[width=0.22\textwidth]{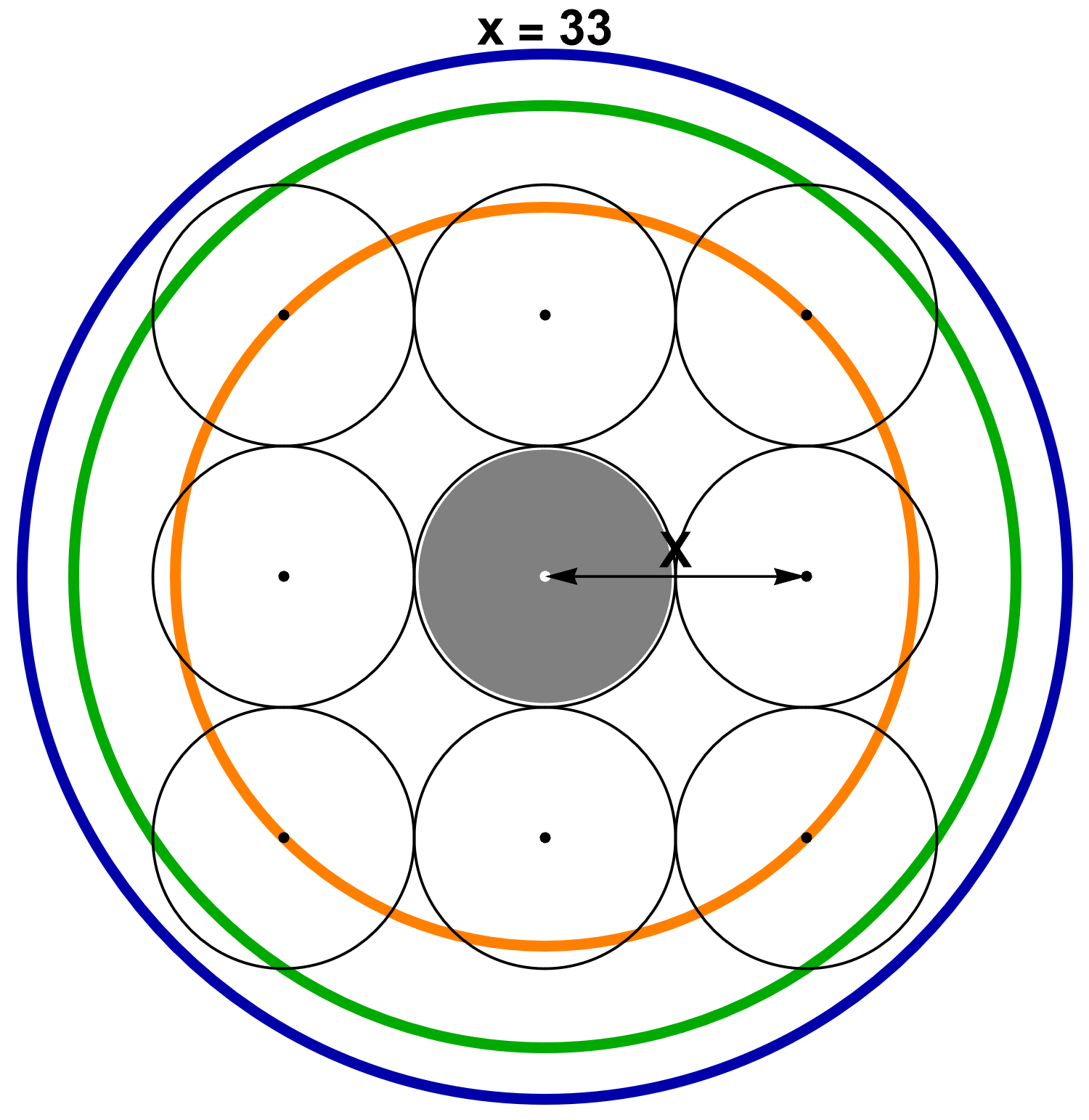}
 & 
  \includegraphics[width=0.22\textwidth]{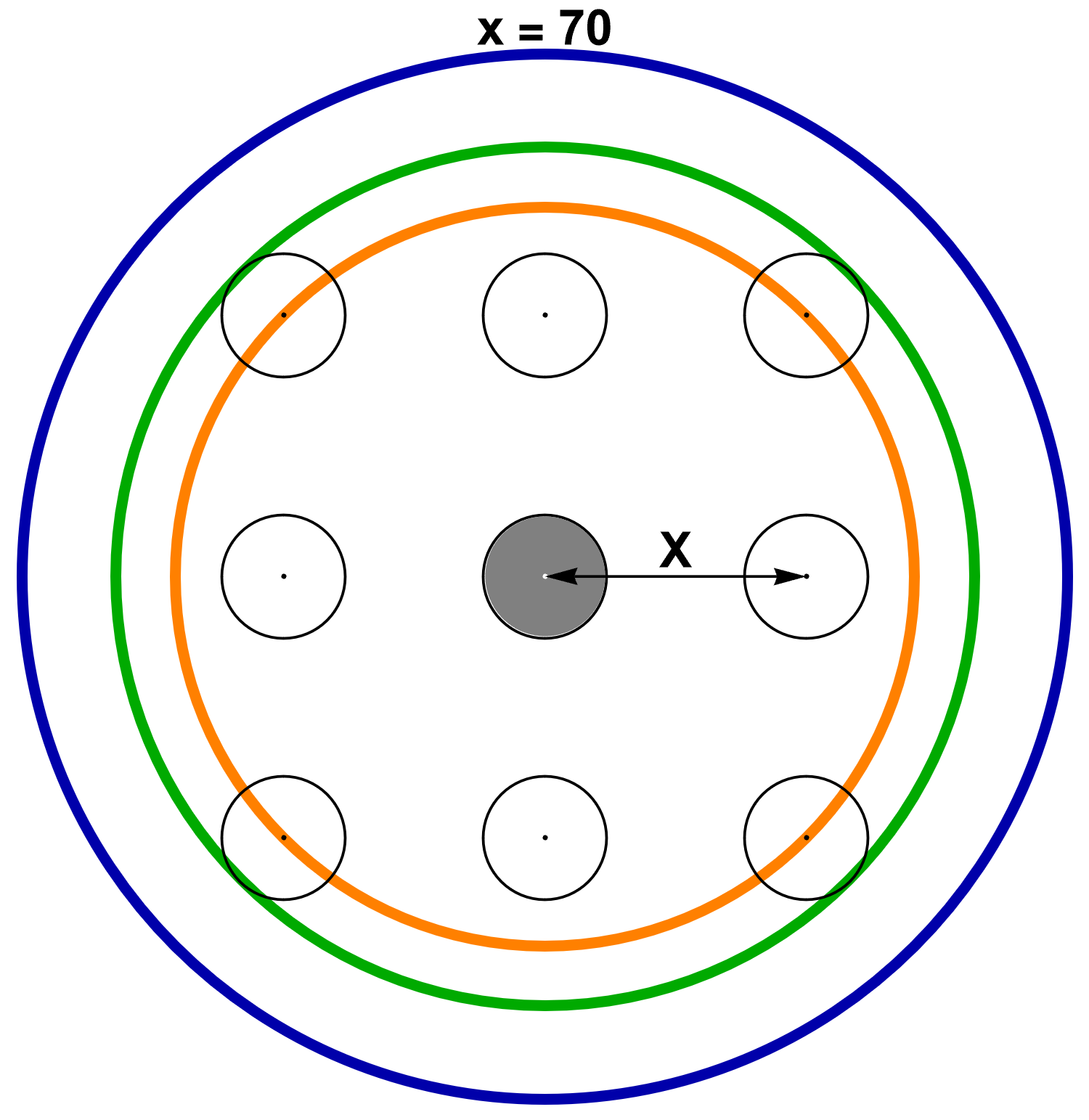}
   & 
  \includegraphics[width=0.22\textwidth]{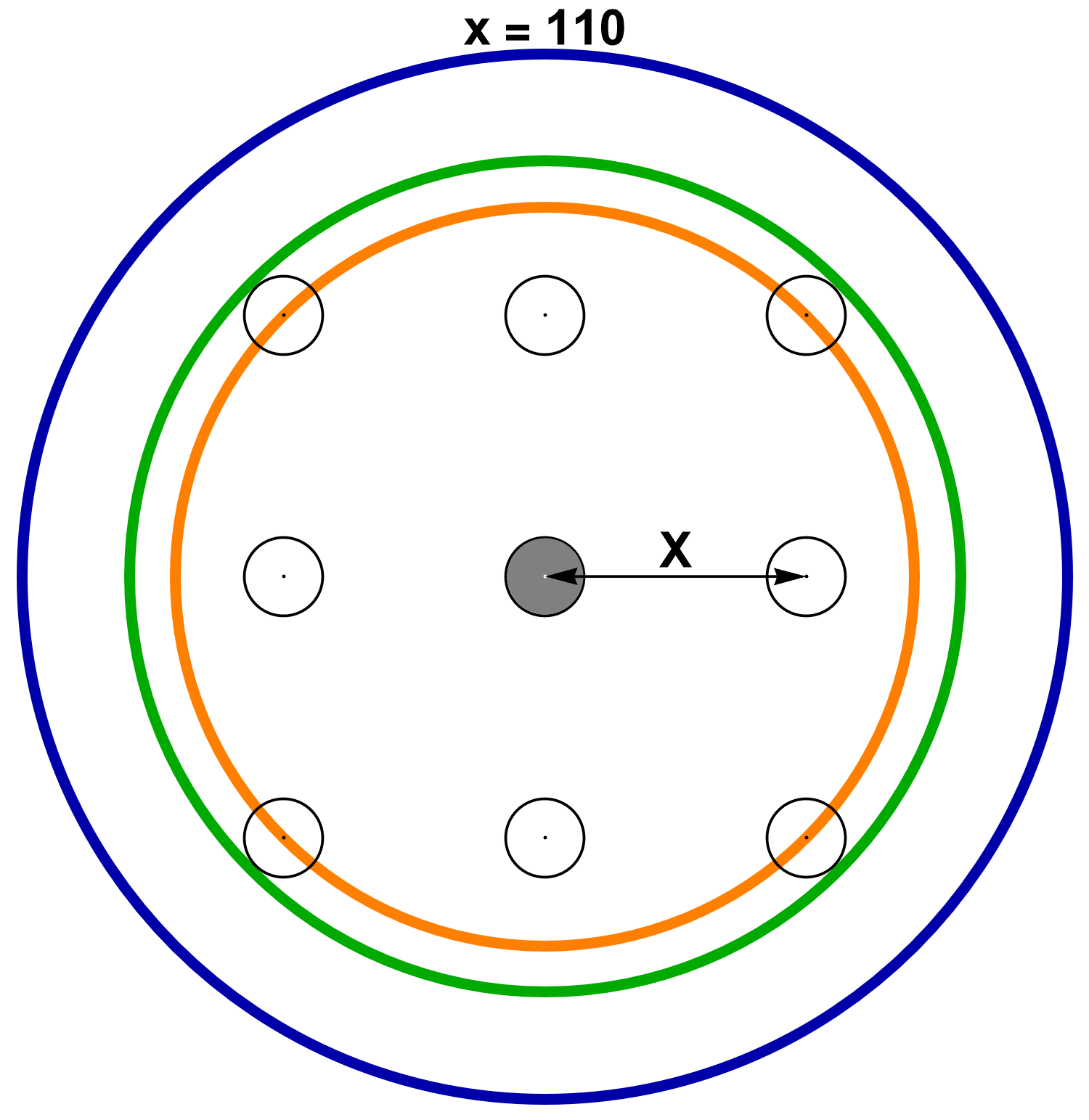}
\\
a) & b) x=33\,mm & c) x=70\,mm & d) x=110\,mm \\
\end{tabular}
\caption{In order to correctly identify the lattice neighbours (e.g. the Moore neighbourhood in a rectangular lattice) each robot filters the incoming messages using the estimated distance of the message's sender (subroutine SR1b). The filtering threshold $r$ must be within two limits, illustrated in the figures by the yellow and blue lines, for the case of noiseless measurements and positions. The limits become more stringent when we consider placement or measurement errors, as discussed in the text. In our implementation, the Kilobots use a simple equation (green line) that lays approximately in the middle of the two limit lines. The panels (b), (c), and (d) show how the focal robot, in the centre, scales its threshold $r$, green line, when inter-robot spacing increases.}
    \label{fig:neighborhood_radius_rings}
\end{figure}

Suitable values of $r$ must also lay between the two extreme values indicated by the same inequalities, i.e. $2x > r > \sqrt{x^2+y^2}$ (and in an analogous way for the equations considering the error $\epsilon$), as also illustrated in Figure \ref{fig:neighborhood_radius_rings}. The Kilobot algorithms benefit from simple computation, thus, in our implementation, the Kilobots compute $r=1.5x + 10$, which is a simple equation that lays approximately in the middle of the two extremes of the inequality and worked consistently well for triangular and rectangular lattices (see Figure~\ref{fig:neighborhood_examples}).

Occasionally, it might happen that due to noise in the transceiver (especially in swarms with small spacing between individual robots), the distance of the sender is consistently overestimated. In such instances, messages sent by a valid neighbour are discarded because the estimated distance is larger than $r$. In this case, robots may create an incomplete list of neighbours. 
Even if a consistent overestimation of distance by a robot is rare, the presence of these errors is relatively high when operating with large swarms because the probability of these rare events occurring increases with the swarm size.
Therefore, we implemented a repairing mechanism as the third phase of the subroutine SR1b, which begins 5\,s after the start of the second phase. During this repairing phase, each robot broadcasts a message with its ID (sender-ID) and all IDs in its neighbour list (neighbours-IDs). Each robot checks all incoming messages (regardless of the sender's distance) and if its ID is one of the neighbours-IDs, it adds the sender-ID to its neighbour list. This approach reduces considerably distance estimation errors which are typically asymmetric, that is only one robot in a couple of neighbours overestimates the distance.




\begin{algorithm}[h!]
		\caption{Subroutines SR1a and SR1b: Locally-unique IDs assignment and neighbour list creation}
		\label{code:SR1b}
    	\begin{algorithmic}[1]
			
				\State Set \textit{$time\_limit_1$ = 300} \Comment{Algorithm parameter}\label{line:timer1} 
			\State Set \textit{$time\_limit_2$ = 800} \Comment{Algorithm parameter}\label{line:timer2}
			\State Set \textit{$time\_limit_3$ = 1600} \Comment{Algorithm parameter}\label{line:timer3}
			\State Set \textit{min\_msg\_distance = 255} \Comment{Minimum distance variable}
			\State Set \textit{i = 1} \Comment{Iterator}
			\State Set \textit{neighbours\_count = 0} 
			\Comment{Initialisation of counter for neighbours}
						\State Set \textit{my\_id = Random()}\Comment{Select random ID from allowed range}
			\State Set \textit{$\overrightarrow{my\_msg}$ = \{my\_id\}} \Comment{Set the content my outgoing message $\overrightarrow{my\_msg}$}

        \ForEach{message $\overrightarrow{m_y}$ received}
            \State Set \textit{received\_id = $m_{y}[1]$} \Comment{Store the ID of my neighbour}
			\State Set \textit{msg\_distance} = $distance\_estimate(m_y$) \Comment{Compute the distance of my neighbour}
			\If{$kilo\_ticks \leq time\_limit_1$}
			    
			   \State Add \textit{received\_id} to \textit{$\overrightarrow{black\_list}$}
			\If{\textit{received\_id} == \textit{my\_id}}\Comment{ 1st phase of SR1a}
			    \State Set \textit{my\_id = random() $\notin$ $\overrightarrow{black\_list}$}\Comment{ Select random ID from allowed range excluding black-listed IDs}
			    \State Set \textit{$\overrightarrow{my\_msg}$ = \{my\_id\}}\Comment{ Update the content of my outgoing message}
			\EndIf
			\EndIf
		
			\If{$kilo\_ticks > time\_limit_1$ AND $kilo\_ticks \leq time\_limit_2$}\Comment{ 2nd phase of SR1a}
			\State Add \textit{received\_id} to \textit{$\overrightarrow{id\_list}$}
				\State Set \textit{$\overrightarrow{my\_msg}$ = \{my\_id,\,$id\_list[i]$\}} \label{line:new-message}
			 
			   \State Set \textit{i++}; \Comment{Iterate throughout the received ids stored in the $\overrightarrow{id\_list}$}\label{line:new-message2}
			\If{$my\_id \in$  $\overrightarrow{m_y}$ } \label{line:duplicated-id-check2}
			     \State Add \textit{my\_id} to \textit{$\overrightarrow{black\_list}$}
			    \State Set \textit{MY\_ID = Random() $\notin$ $\overrightarrow{black\_list}$}
			   
			\EndIf
			\If{\textit{msg\_distance $<$  min\_msg\_distance} AND  \textit{msg\_distance $\ge$ robot\_body\_length} }\label{line:min-distance}
			    \State Set \textit{min\_msg\_distance = msg\_distance}\Comment{Store the minimum distance value}
			   
			\EndIf
			\EndIf
			
    			\If{$kilo\_ticks > time\_limit_2$ AND $kilo\_ticks \leq time\_limit_3$}\Comment{2nd phase of SR1b}
            \If{\textit{received\_id} $\notin$ \textit{neighbours\_list}}\Comment{Avoid double counting}		
              \If{\textit{msg\_distance} $< 1.5 \cdot$ \textit{min\_msg\_distance}+10}\Comment{Distance smaller than $r$}
            		   \State Add \textit{received\_id} to \textit{$\overrightarrow{neighbours\_list}$}\Comment{Store neighbour's ID in neighbours list}
            		   \State Set \textit{neighbours\_count++}
            		  
            		\EndIf
            		\EndIf
    			\EndIf
		    \EndFor
	\end{algorithmic}
\end{algorithm}

\subsubsection{Subroutine SR1c -- Position group identification}
\label{sec:SR1c}

Once the neighbour list is created, subroutine SR1c enables every robot to determine its position group in the lattice (Algorithm \ref{code:SR1c}). Each robot broadcasts messages indicating its ID and the number of neighbours in its list. Once the robots have collected the information regarding the number of neighbours from every neighbour, they use it to determine their own position in the lattice. We consider three possible position groups: CORNER, BORDER, and MIDDLE, as illustrated with colour-coded positions in Figure~\ref{fig:neighborhood_examples}. The robot is positioned on the CORNER of the lattice if it has fewer neighbours than any of its neighbours (line \ref{line:corner} of Algorithm \ref{code:SR1c}). The robot is positioned in the MIDDLE if its number of neighbours is the largest of its neighbourhood (line \ref{line:middle}). The robot is positioned on the BORDER of the lattice in all other cases (line \ref{line:border}, i.e. it has neighbours with a higher number of neighbours and it has not fewer neighbours than any of its neighbours). Note that for a known regular topology, this procedure can be simplified as the position in the formation can be derived directly from the number of identified neighbours (e.g. in a rectangular topology the CORNER has 3 neighbours, the BORDER has 5 neighbours, and MIDDLE has 8 neighbours). However, subroutine SR1c, and more generally routine R1, does not require the robots to know the lattice topology in advance. On the contrary, during SR1c, the robots can self-deduce the regular topology they are part of by using the neighbour counts.

\begin{algorithm}[h!]
		\caption{Subroutine SR1c: Position group identification}
		\label{code:SR1c}
    	\begin{algorithmic}[1]

			\State Set \textit{my\_msg = \{my\_id,neighbours\_count\}}
			\Comment{Set the content of my outgoing message}
				\State Set \textit{max\_count = 0}
				\Comment{Internal variable} \label{line:maxcount}
				\State Set \textit{min\_count = 255}
				\Comment{Internal variable} \label{line:mincount}
 \ForEach{message $\overrightarrow{m_y}$ received}	\State 	Set \textit{received\_id= $m_y$}[1]
			\If{number of neighbours with unknown \textit{neighbours\_count} $>$ 0}
			    
    			\If{$received\_id \in \overrightarrow{neighbours} $ 
    			}
    			\State 	Set \textit{received\_neighbour\_count= $m_y$}[2]
    			\If{\textit{received\_neighbour\_count $<$ min\_count}} 
    			\State Set \textit{min\_count = received\_neighbour\_count }
    			\EndIf
    			\If{\textit{received\_neighbour\_count $>$ max\_count}} 
    			\State Set \textit{max\_count = received\_neighbour\_count }
    			\EndIf
    			\EndIf
            \EndIf
            \If{\textit{neighbours\_count $<$ min\_count}}
            \label{line:corner}
            \State Set \textit{my\_position}=CORNER
            \EndIf
            \If{\textit{neighbours\_count $\ge$ min\_count} AND \textit{neighbours\_count $<$ max\_count}}
            \label{line:border}
            \State Set\textit{ my\_position}=BORDER
            \EndIf
            \If{\textit{neighbours\_count$==$max\_count }}
            \label{line:middle}
            \State Set \textit{my\_position}=MIDDLE
            \EndIf
        \EndFor
		\end{algorithmic}
	\end{algorithm}

			

\subsection{Routine R2 -- Coordinate system construction}

Routine R2 enables the swarm to create a global coordinate system in which every robot self-localises by computing its 2-dimensional coordinates within the lattice. This routine requires the completion of routine R1 by which robots know the locally-unique IDs of their neighbours and know their position group in the lattice (i.e. CORNER, BORDER, or MIDDLE). This routine has been designed for rectangular lattices, however, our subroutines can be potentially extended to different protocols for building global coordinate systems (possibly in dimensions higher than two) for different lattice topologies, e.g. hexagonal \citep{snyder_coordinate_1999}.
For the rectangular lattice, the 2D coordinates are computed with respect to the x and y axes which correspond to two orthogonal edges of the rectangle (that are randomly chosen in subroutine SR2a). The coordinates start from value (1,1) at the origin (a robot in a corner of the lattice) and indicate the discrete positional values for every robot in the lattice. 

\subsubsection{Subroutine SR2a -- Axes origin and direction}
Through subroutine SR2a, one of the four corners is randomly selected as the axes' origin (Algorithm \ref{code:SR2a}).
At the beginning of this subroutine, every CORNER selects a random number, and the CORNER that selected the smallest number is selected as the axes' origin. Because the four CORNER robots are not in direct communication range, all robots cooperate in the communication by spreading the corners' random numbers throughout the lattice. In order to minimise bandwidth usage and speed up the information spreading, every robot only relays the lowest random number that it has received so far and ignores any other higher number (lines \ref{line:relay-corner1}-\ref{line:relay-corner2}).
The CORNER robots also stop broadcasting their random number once they receive a message with a number lower than their own (line \ref{line:emptymessage}). 
After sufficient time, which we estimate as the time necessary to send 25 messages (about 13\,s), we assume that the message representing the lowest random number of one CORNER robot has reached all other CORNERs. The CORNER robot that has not received any lower random number, then, takes on the role of origin of the axes and sets its coordinates to (x,y)=(1,1).

In order to minimise the probability that two random numbers have the same value, the random number should be uniformly drawn from the largest range possible.
For example, in our implementation, the Kilobots can exchange messages with a 9-byte payload (72 bits). Therefore, the CORNER computes its random number in the range $[0,2^{72}]$. Using such a large range, the probability of two identical 9-byte sequences being generated in two CORNER robots is negligible.

Once the axes' origin CORNER robot is selected, it is possible to determine the axes' orientation by assigning coordinates to its neighbours. The axes' orientation is also randomly determined, this time using information already locally available (from SR1a). The CORNER robot assigns to the BORDER node with the lowest locally-unique ID the coordinate (x,y)=(2,1) and to the BORDER node with the highest ID the coordinate (x,y)=(1,2). 
As a result, both the origin and orientation of the coordinate system are random and self-emergent in every run.
This operation terminates subroutine SR2a and enables the start of subroutine SR2b.

\begin{algorithm}[h!]
		\caption{Sub-routine SR2a: Axes origin and direction}
		\label{code:SR2a}
    	\begin{algorithmic}[1]
    	\State Set \textit{$time\_limit_4$ = 400} \Comment{Algorithm parameter} \label{line:timer4}
    	\State Set $\overrightarrow{coord} = \{0,0\}$
    	\Comment{Coordinates pair} \label{line:coord}
    	\State Set \textit{lower\_id\_border = 0}
    	\Comment{Internal variable} \label{line:lib}
    	\State Set \textit{origin = }FALSE
    	\Comment{Internal variable} \label{line:origin}
    	\State Set \textit{start\_time= kilo\_ticks}
    	\Comment{Current value of kilobot internal clock} \label{line:kiloticks}

			\If{my\_position $==$ CORNER}
    			\State Set \textit{origin = }TRUE
       \State Set \textit{my\_random\_num =  Random()}
       \State Set \textit{my\_msg =  my\_random\_num}
       \EndIf
                \ForEach{message $\overrightarrow{m_y}$ received}
                \If{$kilo\_ticks < start\_time + time\_limit_5$}\label{line:SR2a:phase1}
                \If{my\_position $==$ CORNER}
                \If{$m_y[1]<$\textit{my\_random\_num}}
                    \State Set \textit{origin = }FALSE
                    \State Set \textit{my\_msg} = $\emptyset$\label{line:emptymessage}
                \EndIf
            \Else
            

                \If{$m_y[1] <$ min\_received\_IDs[1]}
                \label{line:relay-corner1}
                    \State Set \textit{$\overrightarrow{my\_msg}$ = $\overrightarrow{m_y}$}
                \EndIf \label{line:relay-corner2}
                \EndIf
                
            \Else
            
            \If{origin==1}
                \State Set $\overrightarrow{coord} = \{1,1\}$
                \State Set lower\_id\_border = ID of border node among neighbours with lower ID value. 
                 \State Set \textit{$\overrightarrow{my\_msg}$ = \{my\_id,1,1,lower\_id\_border\}}
            \EndIf
            
            
            \If{my\_position $==$ BORDER AND $m_y[2]==1$ AND $m_y[3]== 1$}
                \If{$m_y[4]$==my\_id}
                    \State Set $\overrightarrow{coord} = \{2,1\}$
                \Else 
                    \State Set $\overrightarrow{coord} = \{1,2\}$
                    
                \EndIf
                \EndIf
            \EndIf
        \EndFor
    
		\end{algorithmic}
	\end{algorithm}

\subsubsection{Subroutine SR2b -- Lattice dimension measurement}
The subroutine SR2b enables the swarm to compute collectively the dimension of the two axes and gives information to the border robots on their position with respect to these axes. Knowing the axes' dimensions also allows every robot to measure the size of the swarm, a quantity that is typically hard to compute with decentralised algorithms \citep{saha_memory_2021, ganesh_peer_2007}.

\begin{figure}[h!]
\centering
  \includegraphics[width=0.95\textwidth]{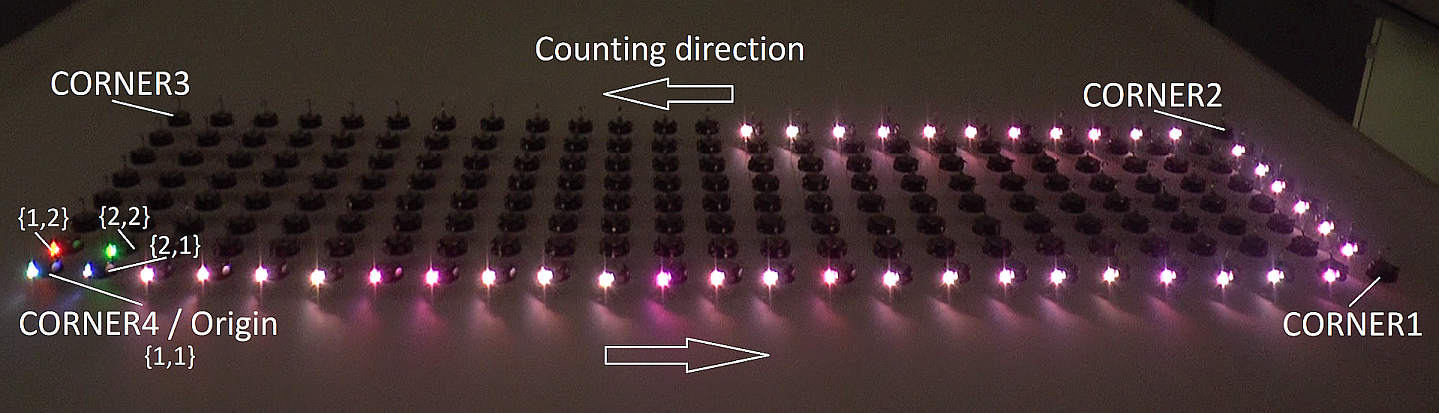}
\caption{Image of the border counting process (subroutine SR2b) in a swarm of 200 Kilobots. The robots on the borders of the $25\times8$ lattice light up when they receive the border-count message during the execution of subroutine SR2b. In this way, it is possible to see the message travelling throughout the lattice edges. In the bottom-left corner, there is the robot at the axes' origin, (x,y)=(1,1), and its neighbours display different colours to signal their coordinates as discussed in subroutine SR2a.}
\label{fig:border-count}     
\end{figure}

Once the axes' origin has been selected (SR2a), the first BORDER robot on the x-axis, that is the robot with coordinates (x,y)=(2,1), starts the subroutine SR2b. The robot initialises a counter variable with the value 2 and sends this value throughout the lattice border; at each message hop, the next border robot increases the count by one and relays the message, as displayed in Figure \ref{fig:border-count}. These messages are flagged with a specific header denoting the border-count subroutine SR2b. These messages are composed of four values (in addition to the header); one value is the border count that is increased at each step, while the other three values are initialised to zero and will be edited by the CORNER robots. The algorithmic implementation for this process is shown in Algorithm \ref{code:SR2b} and indicates that every BORDER robot that did not receive a border-count message yet accepts the message, adds one to the counter, and broadcasts the new value. Three additional rules are necessary for computing correctly the lattice edge dimension. The first additional rule regards BORDER robots that are adjacent to a CORNER robot and received their first message from another BORDER robot. These robots must use a special header in their border-count message. Messages with such special header are only read by CORNER robots so that the other BORDER robot on the diagonal (which did not receive any border-count message yet) will ignore the message and wait for the CORNER robot to send its border-count message with the updated value. The second additional rule requires the CORNER robots to append further information to the border-count message. They include information about their local count that they store in the first zero value composing the message. Therefore the border-count message maintains as separate pieces of information the local count of the three corners traversed during the multi-hop spreading. 
The third additional rule prevents the execution of the counting in both directions. The rule only applies to the robots with coordinates (x,y)=(1,1) and (x,y)=(1,2); these two robots ignore any border-count message with a count value lower than three. This process ends when the count has travelled around the whole border of the lattice; this happens when a valid border-count message is transmitted to the axes' origin. This process is independent of timers and can scale to any grid size larger than or equal to $3\times3$. 
 
Once the origin robot receives the total border count, the total count information and the local counts of the CORNER robots are spread throughout the border in the same direction as before until the message is returned to the origin once again. 
At this point, all BORDER and CORNER robots know the lattice size (total border count and corners' local counts) and their position on the border with respect to the axes' origin (their local count value). The robots can use this information to compute the swarm size and will subsequently use it to compute their coordinates in subroutine SR2c.

\begin{algorithm}[h!]
		\caption{Sub-routine SR2b: Lattice dimensions measurement}
		\label{code:SR2b}
    	\begin{algorithmic}[1]
    	\State Set $my_count = 0$
     \Comment{Local variable for lattice dimension counting} \label{line:mycount}
		 \If{$\overrightarrow{coord} == \{1,1\}$}
		 \State Set $my\_count = 1$
		\EndIf	
		\If{$\overrightarrow{coord} == \{2,1\}$}
		 \State Set $my\_count = 2$
		\EndIf	
		\State Set \textit{$\overrightarrow{my\_msg}$ = \{my\_id,my\_count\}}
	    \ForEach{message $\overrightarrow{m_y}$ received}
	    \If{\textit{my\_position == BORDER AND my\_count == 0 AND $m_y[2]>1$ AND coord $\neq$ \{1,2\}}}
	    \State Set $my\_count = m_y[2]+1$
		\State Set \textit{$\overrightarrow{my\_msg}$ = \{my\_id,my\_count\}}
        \EndIf  
        \If{\textit{coord == \{1,2\} AND my\_count == 0 AND $m_y[2]>3$ }}
	    \State Set $my\_count = m_y[2]+1$
		\State Set \textit{$\overrightarrow{my\_msg}$ = \{my\_id,my\_count\}}

		
	    \EndIf

	    \If{\textit{origin $==$ 1 AND $m_y[2]>3$ }}
		    \State Send sync signal for next phase
		    \EndIf  
	    \EndFor
		\end{algorithmic}
	\end{algorithm}

\subsubsection{Subroutine SR2c -- Coordinates assignment}

Through subroutine SR2c, every robot computes its two coordinates and becomes aware of its position within the full formation. In subroutine SR2c, robots employ the information gathered in SR2b and incrementally assign coordinates, commencing from the edges of the lattice and moving inwards.

\paragraph{Corner and borders.} The corner at the origin already has its coordinates, assigned during subroutine SR2a, (x,y)=(1,1). The other three CORNER and all the BORDER robots use the information exchanged in the border-count messages in SR2b to set their coordinates. The border-count messages contain ordered information on the local count of the three corners encountered during the multi-hop count on the border. We label as C1, C2, and C3, the local counts of the first, second, and third corners encountered during the border-count process (see Figure \ref{fig:border-count}). The CORNERs and the BORDERs  then assign their coordinates through Algorithm \ref{code:SR2aCB}, where $my\_count$ is the local count of the robot running the algorithm and \textit{coord} its coordinates.


\begin{algorithm}[H]
		\caption{Part of subroutine SR2c: Coordinates assignment for CORNER and BORDER robots}
		\label{code:SR2aCB}
    	\begin{algorithmic}[1]

    \If{$my\_position == CORNER$ OR $my\_position == BORDER$}

        \If{$my\_count \leq C1 $}
        \State Set $coord = \{my\_count,1\}$
		\EndIf	

        \If{$my\_count > C1$ \textit{AND} $my\_count \leq C2 $}
        \State Set $coord = \{C1,my\_count-C1+1\}$
		\EndIf	
  
        \If{$my\_count > C2$ \textit{AND} $my\_count \leq C3 $}
        \State Set $coord = \{C1+C2-my\_count,C2-C1+1\}$
		\EndIf	

        \If{$my\_count > C3 $}
        \State Set $coord = \{1,C2+C3-C1-my\_count+1\}$
		\EndIf	
    \EndIf	
		\end{algorithmic}
	\end{algorithm}

\paragraph{Middle robots.}
The MIDDLE robots---which are not on the edge of the lattice---assign their two coordinates independently using information locally broadcast by their neighbours. Once any robot self-assigns its coordinates, it broadcasts its values to its neighbours. Each robot locally stores the coordinates of its neighbours and once it receives the coordinates with three consecutive values on one axis, it self-assigns the middle value on that axis. For instance, a robot $i$ that receives the coordinates from neighbours $j,k,$ and $w$ with values $(x_j,y_j)=(3,7)$, $(x_k,y_k)=(4,7)$, and $(x_w,y_w)=(5,7)$, will set its coordinate $x=4$. Similarly, when a robot $i$ receives the coordinates, $(x_j,y_j)=(3,7)$, $(x_k,y_k)=(3,8)$, and $(x_w,y_w)=(5,9)$, will set its coordinate $y=8$. This process allows every robot to self-assign its coordinates and to become aware of both its position within the lattice and the relative position of all its neighbours. This information can be employed for various subsequent tasks which exploit spatial awareness of the robots (e.g. dynamic role assignment as a function of the robots' position \citep{pratissoli_soft-bodied_2019}, or exploit the swarm-level agreement obtained with routine R2 (e.g. use the coordinates to assign globally-unique IDs).
 
\subsection{Routine R3 -  Synchronised dynamic role assignment}
Routine R3 consists in assigning a different role, or task, to each robot depending on its coordinates. This operation is achieved by providing all robots with the desired action plan. The plan indicates what is the role of every robot depending on its coordinates and how the roles change over time. While all the robots know the full action plan in advance, each robot only knows its role once it computes its coordinates through routine R2. Using the synchronisation method described in Section \ref{seC:synch}, the robots can also synchronously move to the next step of the action plan. Such a synchronous role change can be interpreted as a change in the swarm state. Enabling swarm state changes allows the programming of robot swarms through swarm-level finite state machines, which can simplify the design of collective behaviour, as was also suggested by previous research \citep{pinciroli_swarm-oriented_2016}. Indeed, designing swarm robotics algorithms is complicated as the collective behaviour must be encoded in individual robot rules and the link between swarm and robots can be counter-intuitive. Having the possibility of defining the swarm action plan, while maintaining a fully decentralised approach, can be useful.

In this study, we implemented two types of action plans 
that differ in how the robots activate.
In both action plans, a subset of robots in predefined locations becomes active. In the first case, the active robots activate their motors and move out from the formation. In the second case, the active robots light on their coloured LED and create a collective pattern as shown in Section \ref{sec:experiments}. In the second action plan, robot activation is periodically alternated between a cycle of robot subsets with potentially different active roles (implemented as different light colours). The repeated patterns allowed us to test the robustness of the system (e.g., to maintain swarm-level synchronisation) during several-minute-long runs.


\section{Experiments}
\label{sec:experiments}

We tested the algorithms through a series of experiments in simulation and with swarms of real robots. As indicated earlier, the chosen robotic platform is the Kilobot robot \citep{rubenstein_kilobot_2012,rubenstein_kilobot_2014}, which is a relatively simple robot that can move using vibration motors and exchange IR messages with other robots in a range of about 10\,cm. The Kilobots can also estimate the distance (but not the position) of the sender of any received message. Finally, they can show their internal state through a coloured light-emitting diode (LED).

The code to run our algorithm on the Kilobots is open source and available on Github at \url{https://github.com/TBU-AILab/Kilobot-self-localization}.

\subsection{Simulations}
The simulations have been performed with ARGoS \citep{pinciroli_argos_2012}, a modular and efficient simulator for swarm robotics \citep{pitonakova_feature_2018}, that offers a convenient plugin for simulating the Kilobots \citep{pinciroli_simulating_2018}. The Kilobot-plugin for ARGoS allows the user to implement robot code that can be transferred without any change from simulation to the robots, therefore  easing implementation and testing.

\begin{figure}[htbp]
\begin{tabular}{cc}
  \includegraphics[width=0.45\textwidth]{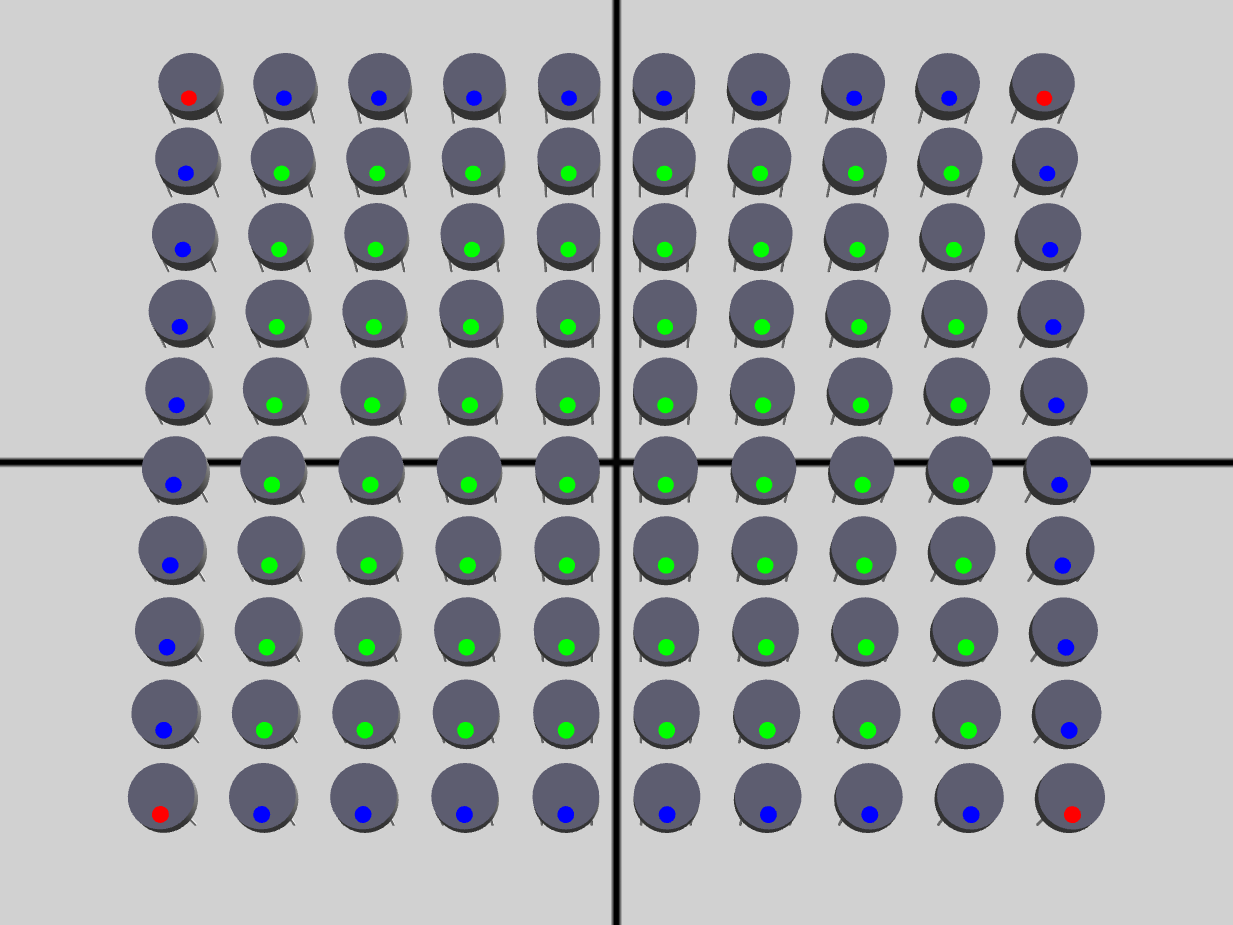}
 & 
  \includegraphics[width=0.45\textwidth]{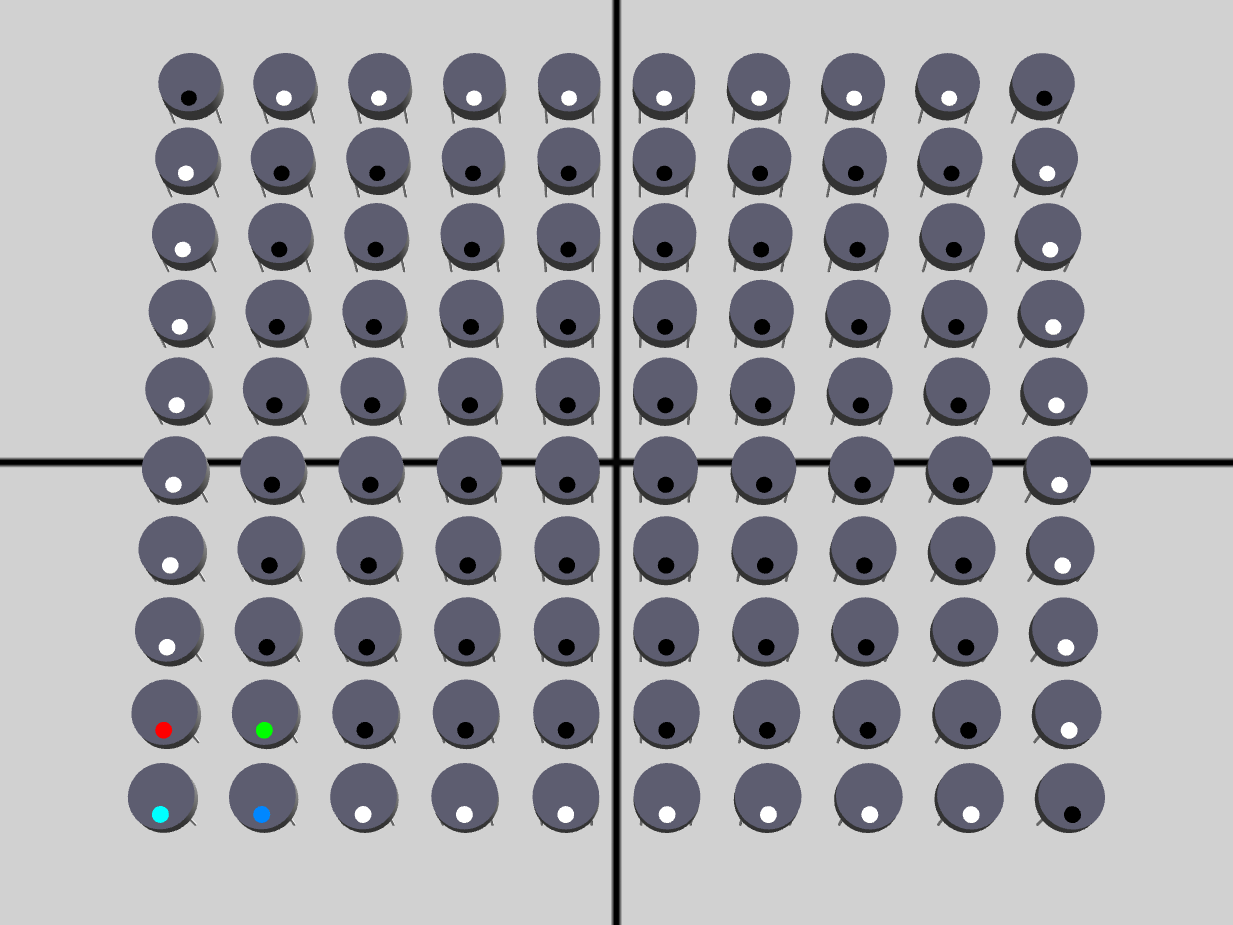}

\\

a) Position group identified (SR1c) & b) Lattice dimension measured (SR2b) \\

  \includegraphics[width=0.45\textwidth]{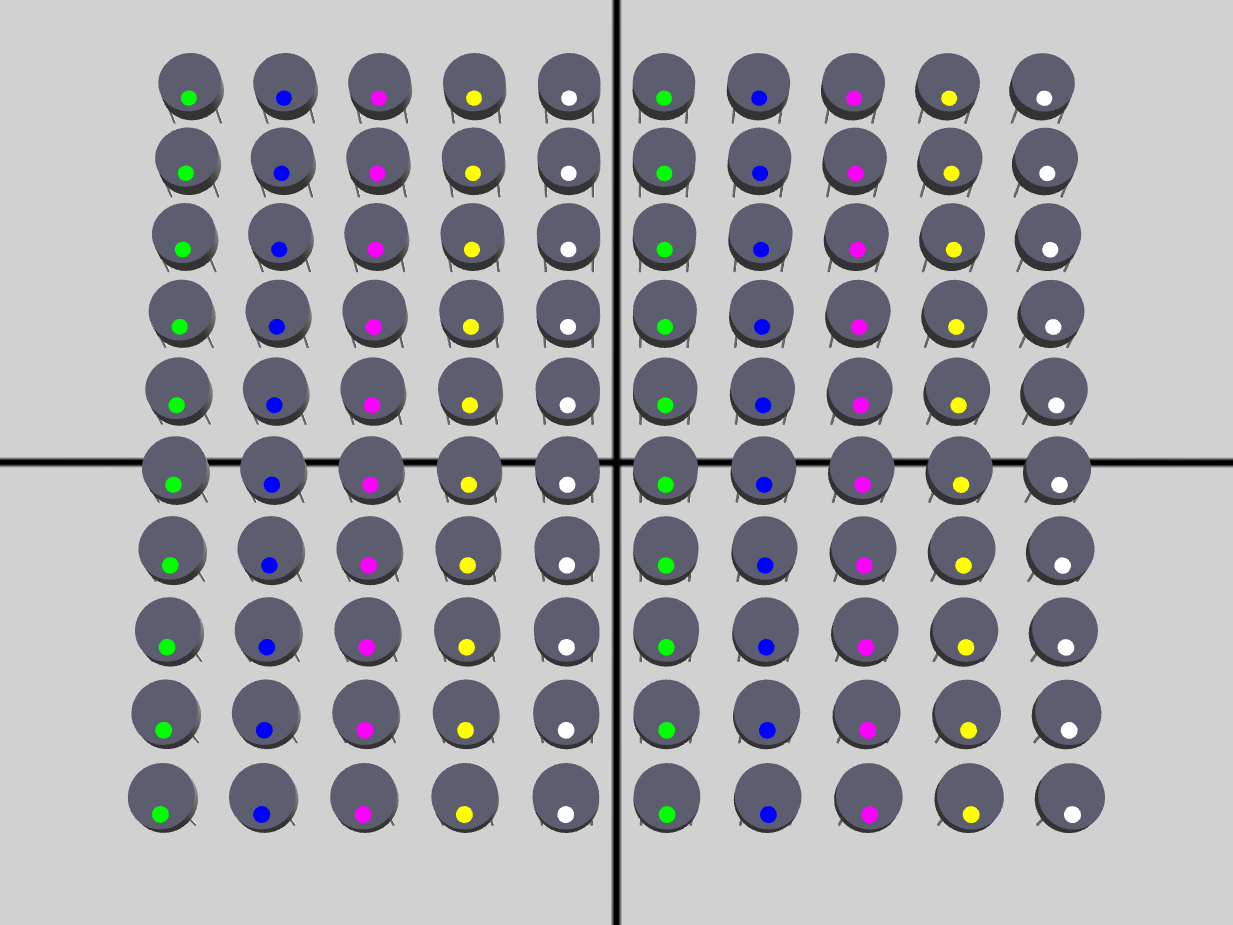}
 & 
  \includegraphics[width=0.45\textwidth]{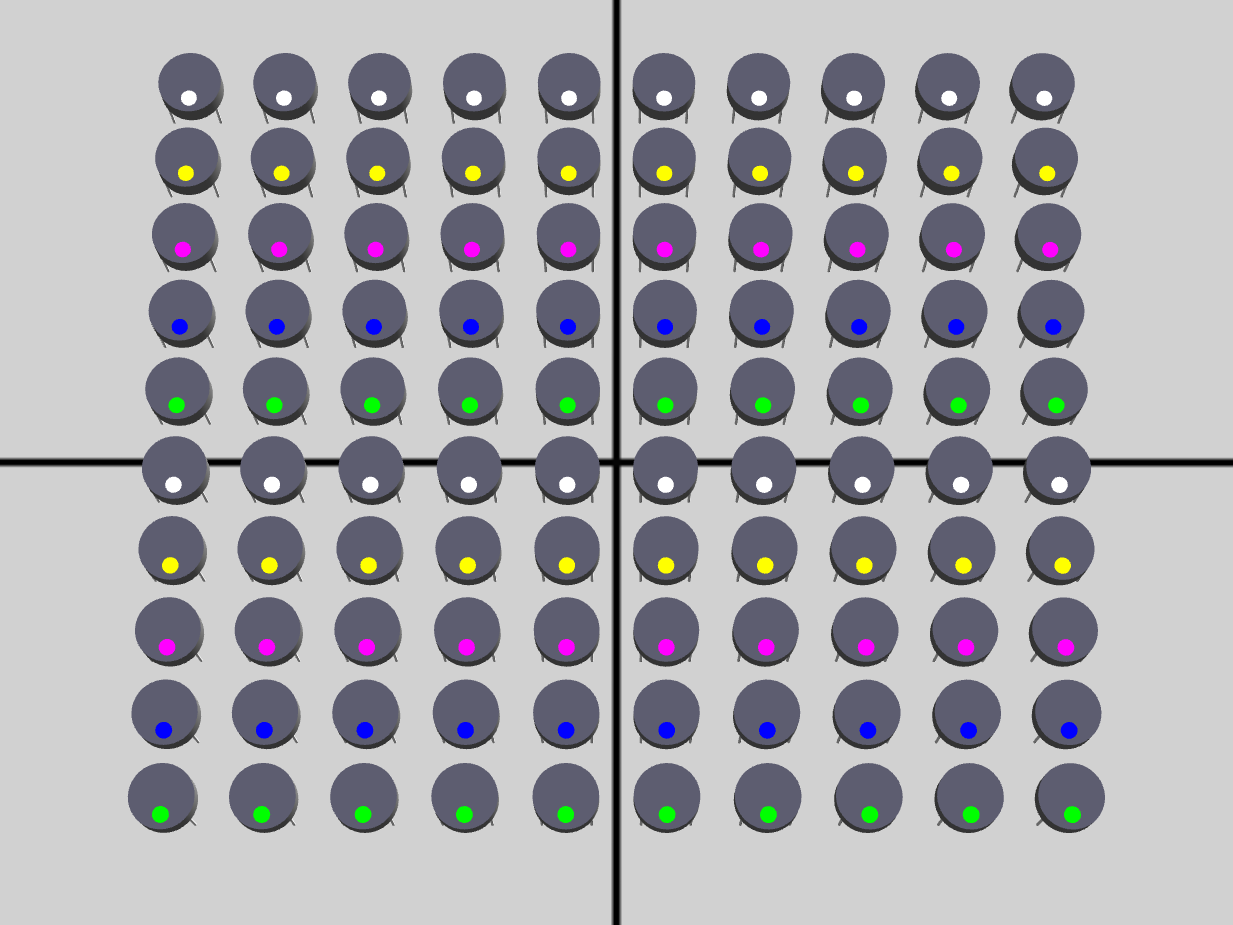}

\\

c) Colour display based on x-axis coordinates & d) Colour display based on y-axis coordinates\\
\end{tabular}
\caption{Four key screenshots from an ARGoS simulation with 100 Kilobots. Panel (a) shows the completion of subroutines SR1c where robots in the three position groups CORNER, BORDER, and MIDDLE light up with colours red, blue, and green, respectively; thus, reproducing the results obtained with real Kilobots in Figure \ref{fig:neighborhood_examples}a. Panel (b) shows the completion of subroutine SR2b with the axes' origin in the bottom left corner, and the robots on the borders that have received the border-count message have their white light on, analogous to Figure \ref{fig:border-count}. In panels (c) and (d), the robots self-assigned roles (routine R3) depending on their x-axis and y-axis coordinates, respectively. The coordinate-based colours are only based on fives colours, which therefore are repeated twice in the $10\times5$ lattice.}
    \label{fig:ARGoS}
\end{figure}

Through simulation, we performed several tests, running numerous independent repetitions (using different random seeds) for each investigated condition. We conducted tests in experiments with up to 1000 simulated Kilobots, where we tested rectangular lattices of different dimensions, from $3\times3$ to $40\times25$, and in each case, we varied the inter-robot spacing, from 35\,mm to 70\,mm. In all our experiments, the swarm successfully completed the process: every robot correctly computed its coordinates and took on its expected role.
Figure \ref{fig:ARGoS} shows four screenshots of a simulation with $10\times10$ Kilobots illustrating the four main phases of the algorithm: neighbourhood construction (Fig. \ref{fig:ARGoS}a), axes' origin selection and border counting (Fig. \ref{fig:ARGoS}b), and assignment of the x-axis values (Fig. \ref{fig:ARGoS}c) and of the y-axis values (Fig. \ref{fig:ARGoS}d).

\subsection{Kilobot experiments}

\begin{figure}[h!]
  \includegraphics[width=0.24\textwidth]{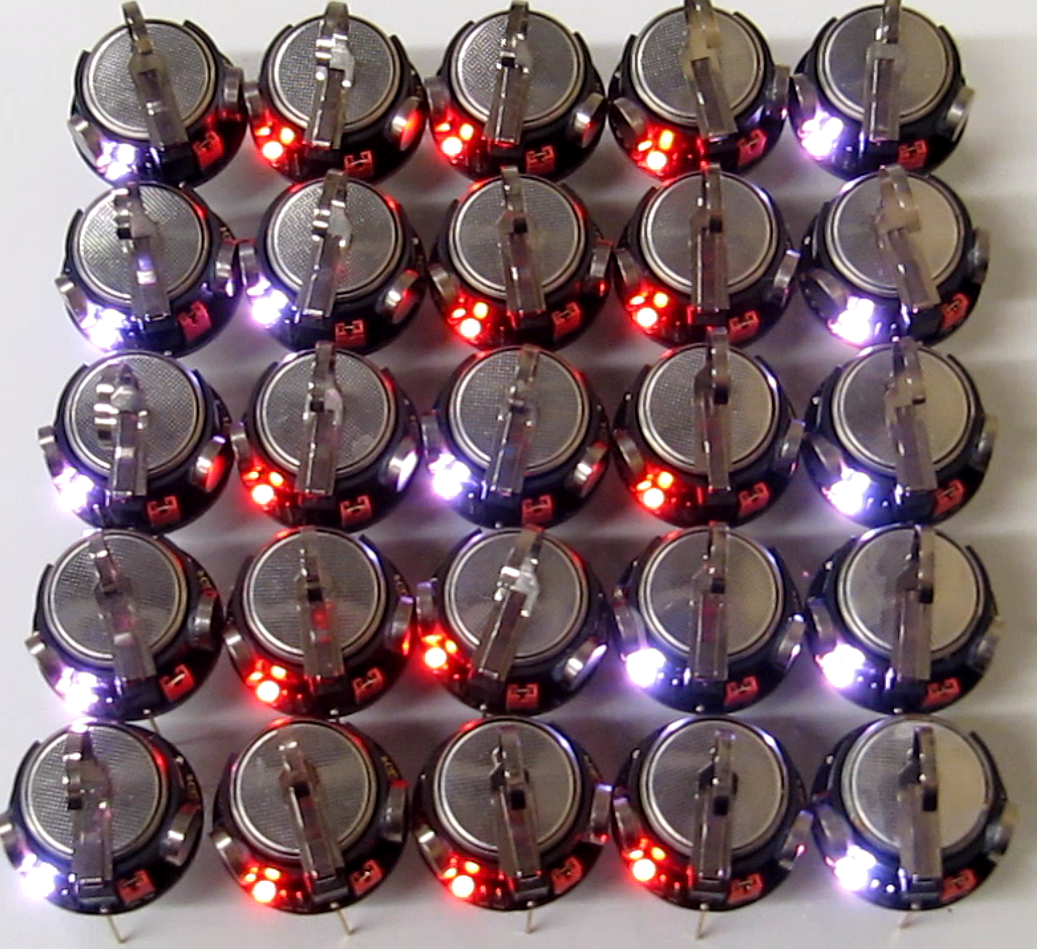}
  \includegraphics[width=0.24\textwidth]{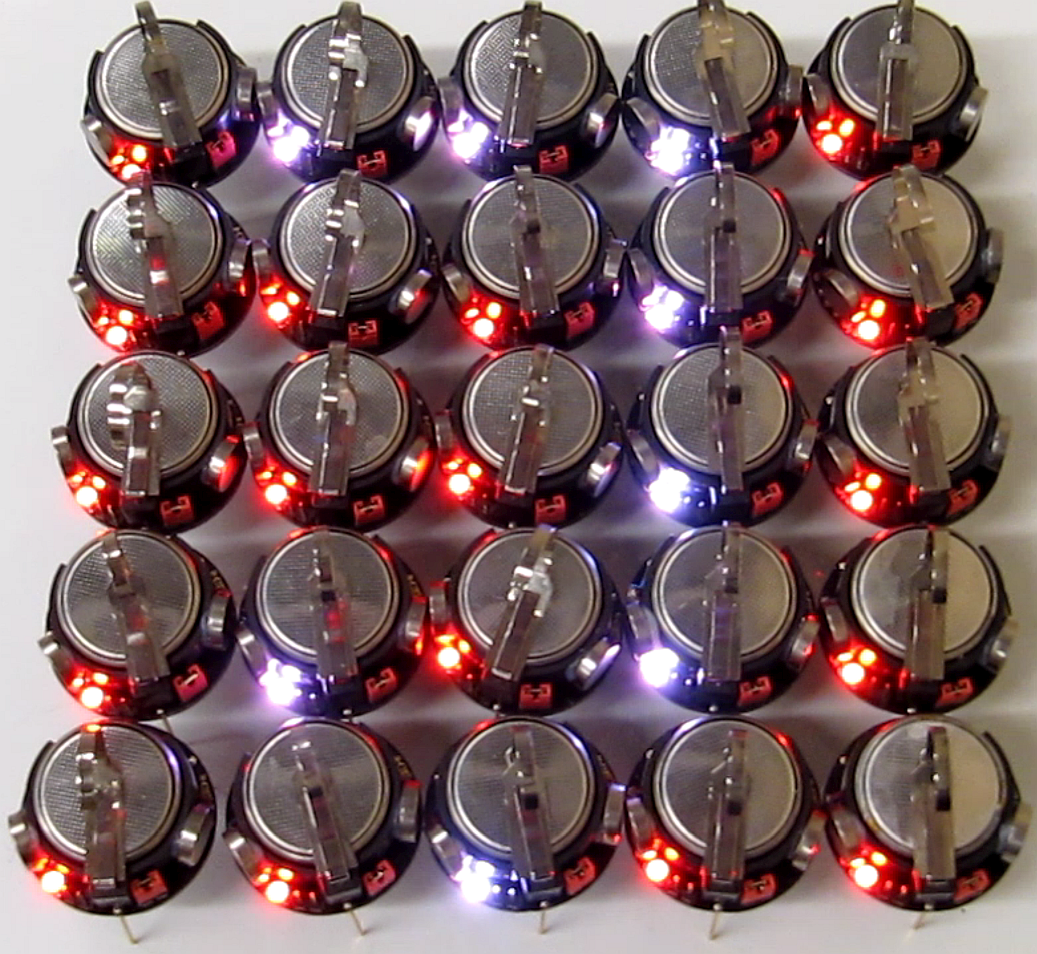}
  \includegraphics[width=0.24\textwidth]{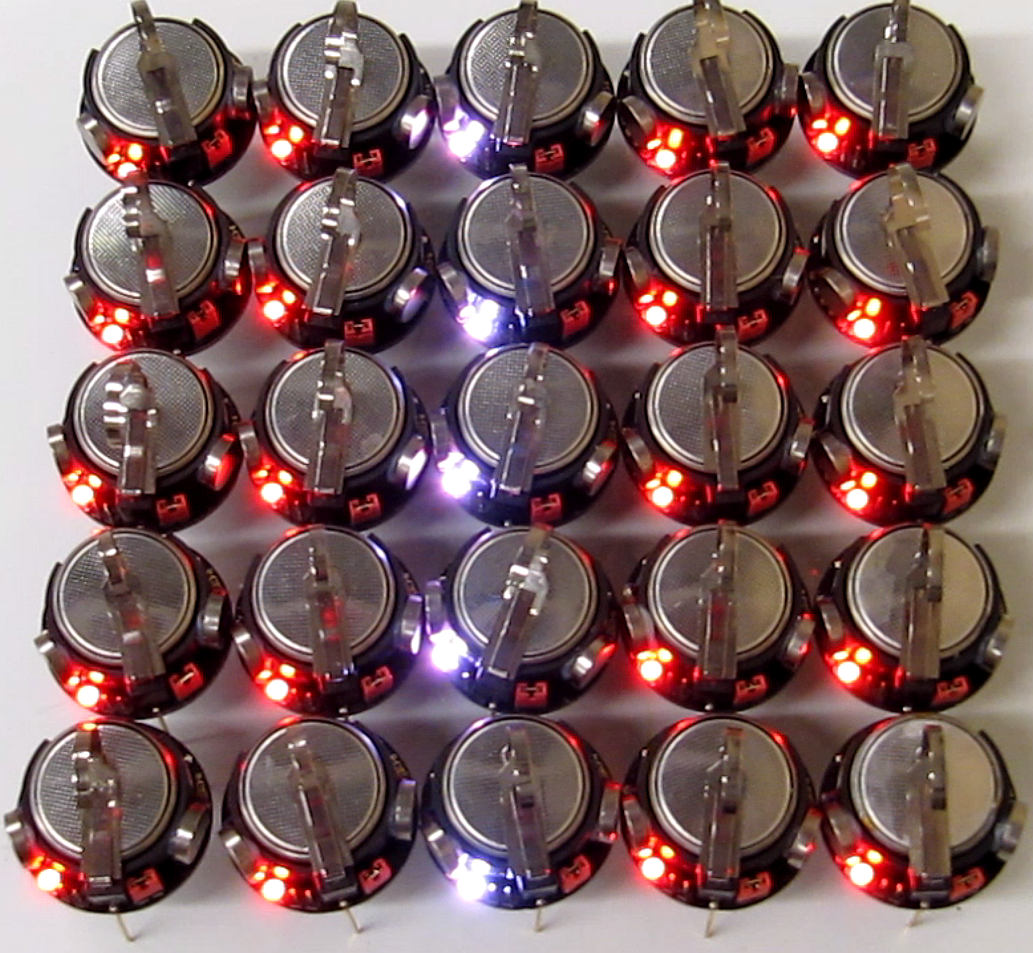}
  \includegraphics[width=0.24\textwidth]{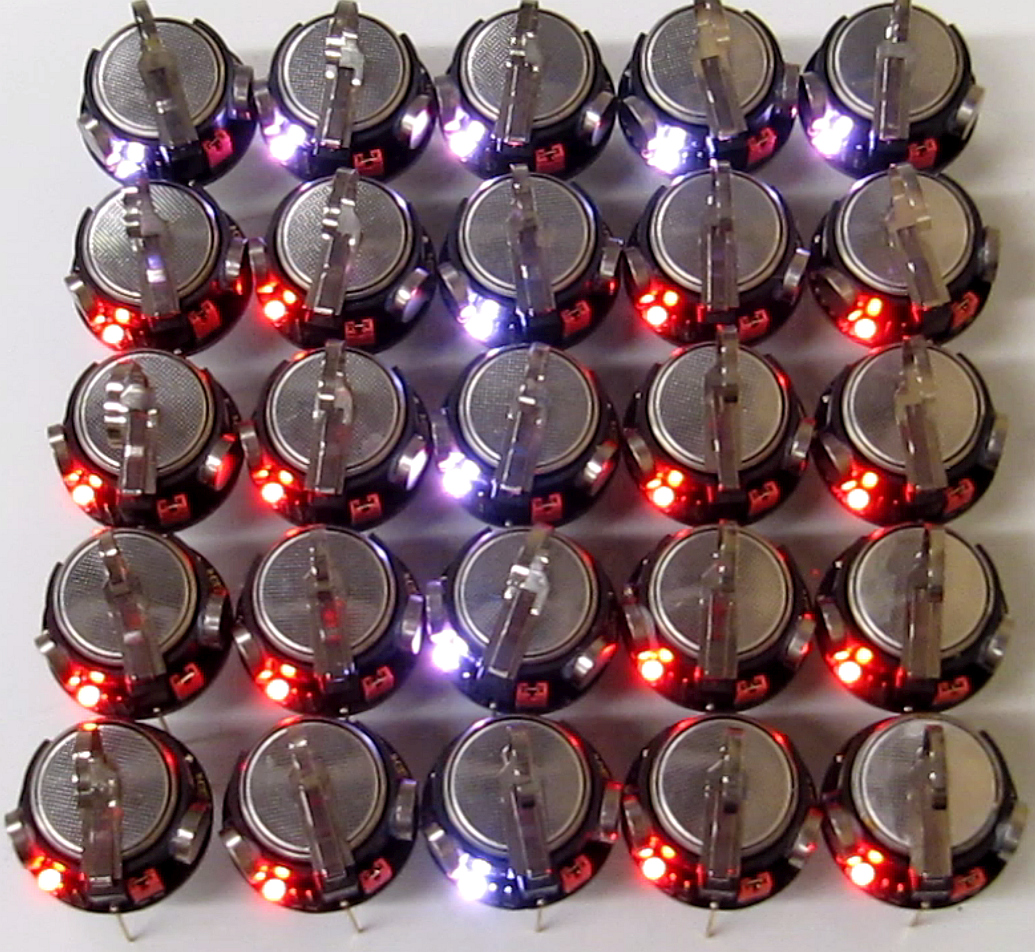}
\caption{
25 Kilobots in a square $5\times5$ lattice self-assign roles in synchrony with each other. They form a repetitive cycle of light patterns displaying the letters N-J-I-T.}
    \label{fig:pattern_NJIT}
\end{figure}

\begin{figure}[h!]
\begin{tabular}{cc}
  \includegraphics[width=0.45\textwidth]{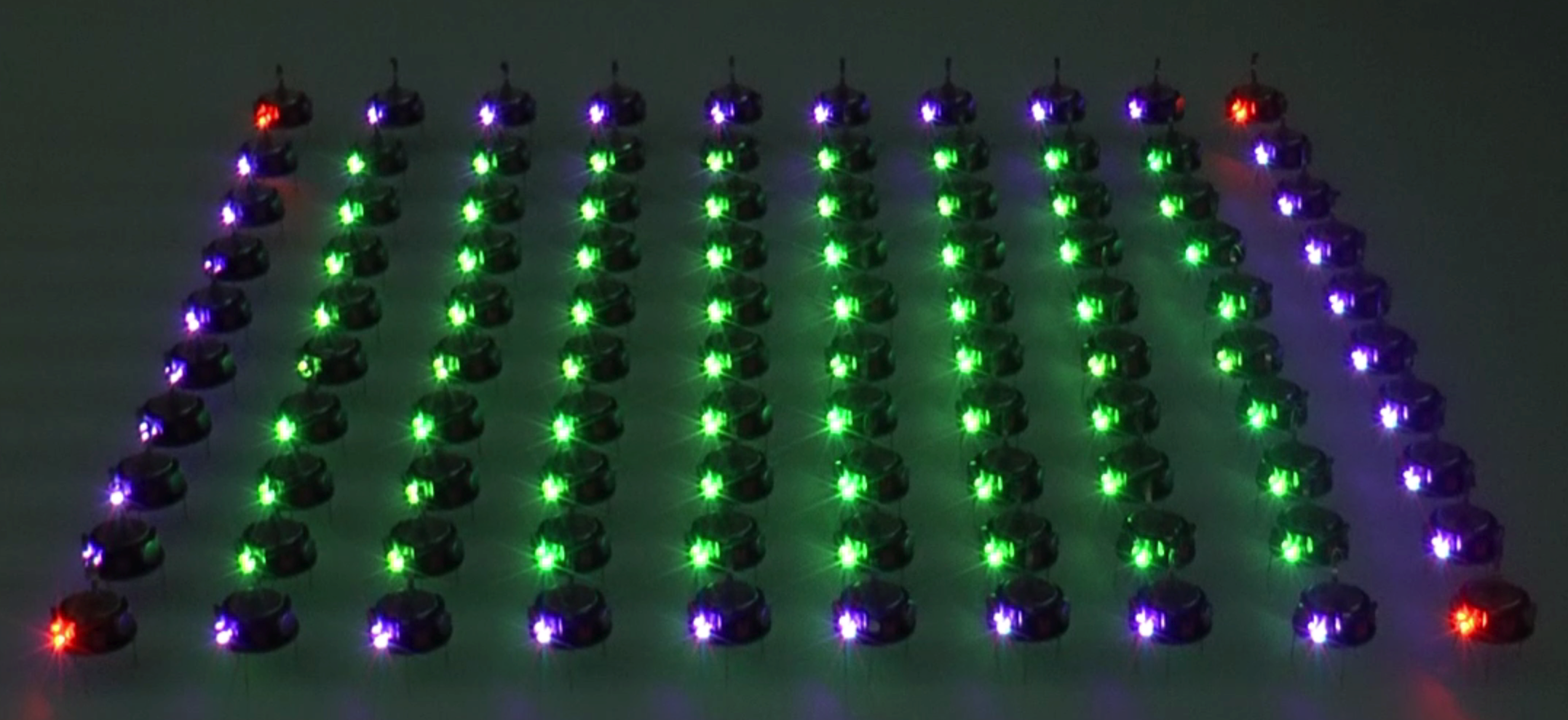}
 & 
  \includegraphics[width=0.45\textwidth]{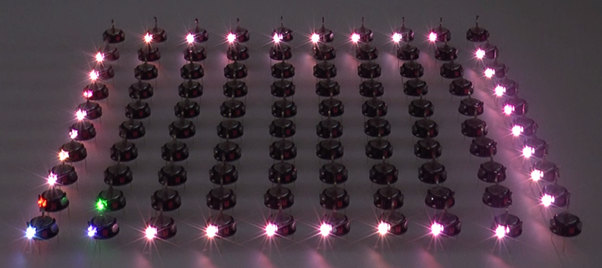}

\\

a) & b) \\

  \includegraphics[width=0.45\textwidth]{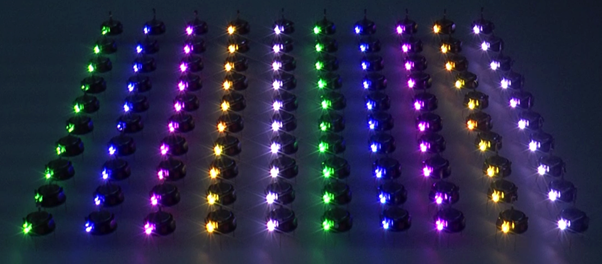}
 & 
  \includegraphics[width=0.45\textwidth]{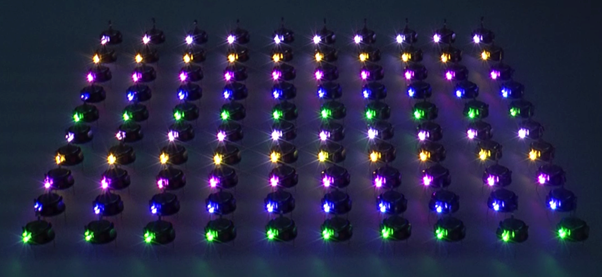}

\\

c) & d)\\

  \includegraphics[width=0.45\textwidth]{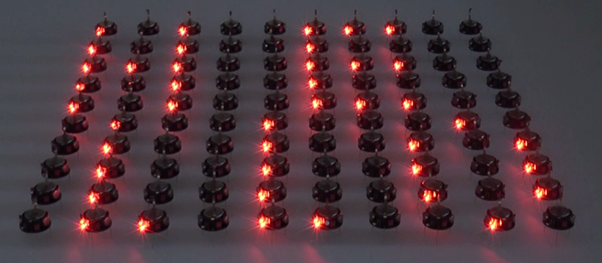}
 & 
  \includegraphics[width=0.45\textwidth]{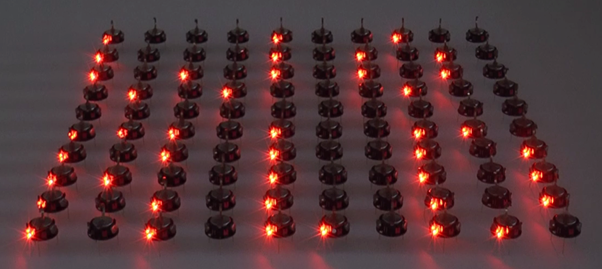}

\\

e) & f)\\
\end{tabular}
\caption{Six screenshots of an experiment with 100 Kilobots in a square $10\times10$ lattice. Panels (a)-(d) display the same four moments as displayed in Figure \ref{fig:ARGoS} with simulated robots. Panels (e) and (f) show two moments of the synchronised dynamic role assignment (routine R3), where the robots form a repetitive cycle of light patterns displaying the words "HE-LLO" and "WO-RLD", respectively. The video of the full experiment is available at \url{https://youtu.be/KlooXOOvZsY}.}

    \label{fig:Sheffield_100}
\end{figure}

\begin{figure}[h!]
\begin{tabular}{c}
  \includegraphics[width=0.85\textwidth]{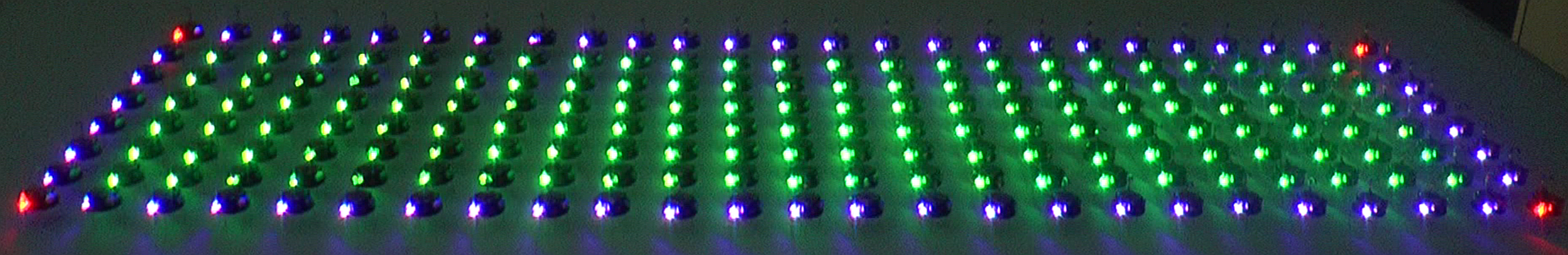}
 \\ 
 a) \\
  \includegraphics[width=0.85\textwidth]{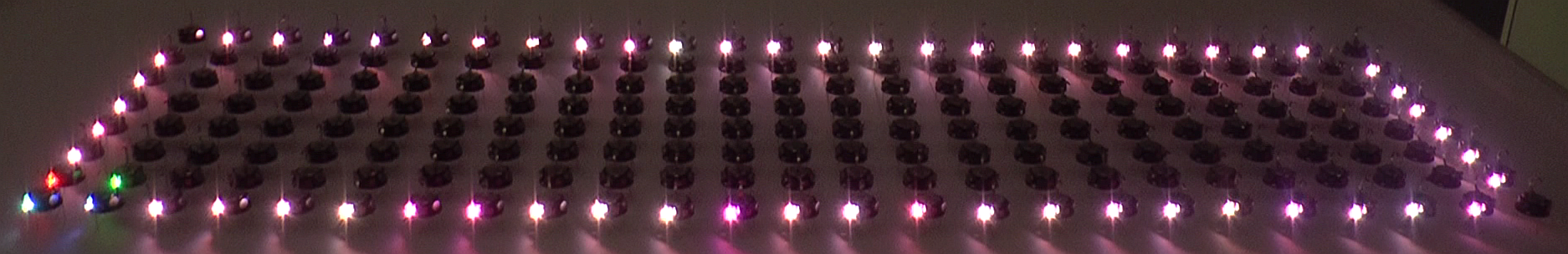}

\\

 b) \\

  \includegraphics[width=0.85\textwidth]{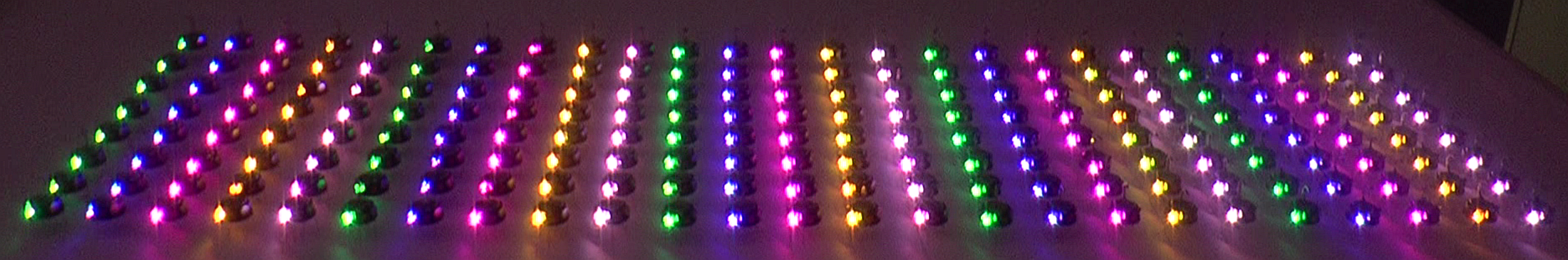}
 \\ 
 c) \\
 
  \includegraphics[width=0.85\textwidth]{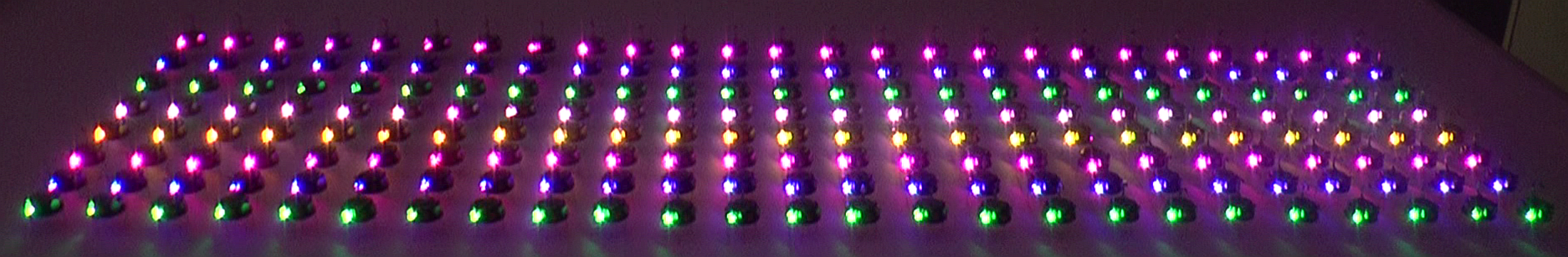}

\\

 d)\\

  \includegraphics[width=0.85\textwidth]{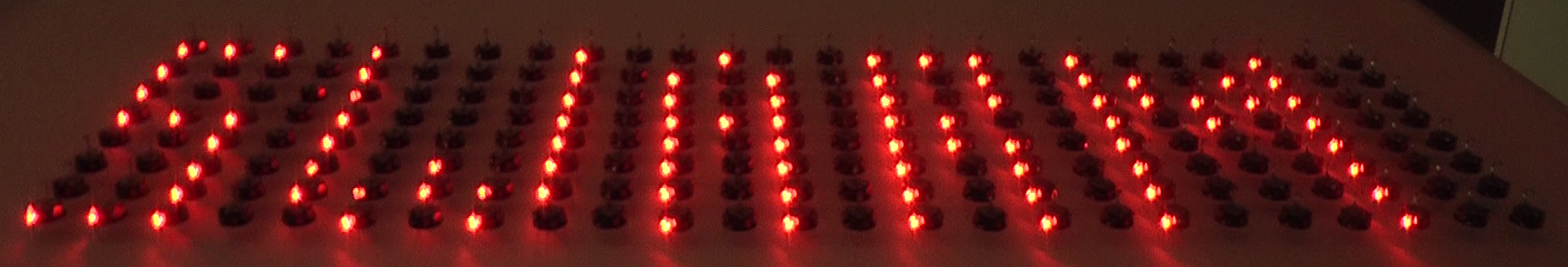}
 \\ 
 e) \\
  \includegraphics[width=0.85\textwidth]{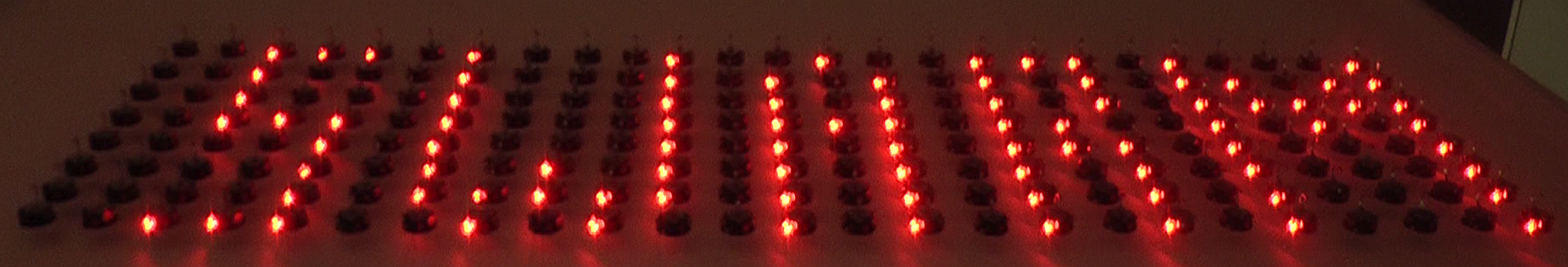}

\\

 f)\\
\end{tabular}
\caption{Six screenshots of an experiment with 200 Kilobots. Panels (a)-(d) display the same four moments as displayed in Figures \ref{fig:ARGoS} and \ref{fig:Sheffield_100} in smaller groups. Panels (e) and (f) 
show two moments of the synchronised dynamic role assignment (routine R3), where the robots form a repetitive cycle of light patterns displaying the word ``SWARM'' shifted by two columns. The video of the full experiment is available at \url{https://youtu.be/S4s6fpWZvMM}.}
    \label{fig:Sheffield_200}
\end{figure}

We also conducted a range of real-robot experiments employing up to 200 Kilobots. We tested different grid sizes, topologies, and inter-robot spacing, as illustrated in Figures \ref{fig:neighborhood_examples}, \ref{fig:pattern_NJIT}, \ref{fig:Sheffield_100}, and \ref{fig:Sheffield_200}.
Small-scale experiments, with up to 25 Kilobots, have been conducted in the Swarm Lab at the New Jersey Institute of Technology, while large-scale experiments with 64 to 200 robots have been conducted using the Kilobot infrastructure of Sheffield Robotics \citep{nikolaidis_characterisation_2017} at the University of Sheffield.

\paragraph{Scalability.} Kilobot experiments tested the capability of the algorithm to run without change in swarms of different sizes. We conducted experiments with grids of $5\times5$ robots (Fig. \ref{fig:pattern_NJIT}), $8\times8$ robots, $10\times10$ robots (Fig. \ref{fig:Sheffield_100}), and $25\times8$ robots (Fig. \ref{fig:Sheffield_200}). The figures show key frames of the process, always terminating with the dynamic role assignment in the form of light patterns that show written text. Note that the axes' origin and orientation are determined at run time and are randomly chosen, therefore in about half of our experiments the text appeared mirrored (when the axes' orientation is inverted the displayed text is subject to a reflection). For ease of visualisation, we only report images and videos of runs in which the light pattern is displayed in the same orientation as the observer.

In Figure \ref{fig:pattern_NJIT}, the 25 robots have a cyclic role assignment in which they form the light pattern N-J-I-T. The video of one experiment with 25 robots is available at \url{https://youtu.be/RdUs_EHRnUU}. Figure \ref{fig:Sheffield_100} shows frames for a swarm of 100 Kilobots in some of the key algorithm phases which terminate with the cyclic role assignment of two light patterns forming the worlds "HE-LLO" and "WO-RLD". The video of one of the experiments with 100 robots is available at \url{https://youtu.be/KlooXOOvZsY}. Finally, Figure \ref{fig:Sheffield_200} shows frames from experiments with the largest swarm tested, organised in a rectangular lattice sized $25\times8$. In these experiments, the cyclic role assignment displays the word "SWARM" that alternates between two different positions in the lattice. The video of one of the experiments with 200 robots is available at \url{https://youtu.be/S4s6fpWZvMM}. Our three experiments with 200 robots took on average 3 minutes to reach completion of routine R2 (where all robots successfully self-assigned their coordinates). While the speed of the process can be potentially optimised by tailoring the time limits of the subroutines to the specific scenario, our tests using the time limits indicated in Section \ref{sec:method} already show a relatively quick process compared with other research experiments that used large-scale swarms of Kilobots \citep{rubenstein_programmable_2014,reina_ark_2017,reina_effects_2018,gauci_programmable_2018}.



\paragraph{Lattice topologies and inter-robot spacing.} 
Through a set of Kilobot experiments, we tested different topologies and spacing among robots. While routine R1 can work on a variety of different regular lattices (as discussed in Sec.\ref{sec:routineR1}), routine R2 has been designed for rectangular lattices only. Figure \ref{fig:neighborhood_examples} shows the final stage of routine R1 in the square and hexagonal lattices, in which the robots display with distinct colours their position group (i.e. CORNER, BORDER, or MIDDLE). Figures \ref{fig:pattern_NJIT} and \ref{fig:Sheffield_100} show square formation with largely different spacing among robots. Finally, in the experiments of Figure \ref{fig:Sheffield_200}, we tested a rectangular lattice (non-squared). In addition, these tests also validate the robustness of our algorithm to misplacement errors which have been introduced by the manual placing of the robots in the lattice formation. For example, in Figure \ref{fig:Sheffield_100} it is possible to appreciate the visible misalignment of some of the robots on the second column from the right, which are particularly evident when the robots have their lights turned on.

\paragraph{Dynamic role assignment.}
In Figures \ref{fig:pattern_NJIT}, \ref{fig:Sheffield_100}, and \ref{fig:Sheffield_200}, the role assignment consisted in displaying a repetitive cycle of light patterns. We let the swarm run these light patterns for several dozen minutes and in all cases the swarm never desynchronised. In an additional experiment, we tested a different role assignment scenario in which a subset of Kilobots at given locations commenced movement and left the formation. The video of this experiment is available at
\url{https://youtu.be/4wlstUNpNcU}.

\paragraph{Summary.} The robot experiments confirmed the validity of the proposed algorithm by consistently reaching the successful completion of all routines. In our scalability tests, we showed that the same identical algorithm can be employed for small grids as well for large-scale experiments, in our case up to 200 real robots. The final routine of cyclic role assignment has also been instrumental to validate the possibility of changing the collective state of the swarm in a synchronised way, even in large swarms composed of robots that only use local communication.
Additionally, real robot experiments tested the ability of the algorithm to operate under variable levels of communication noise and distance measurement errors, intrinsic to such simple devices. Finally, tests with different topologies and with robot formations with some placement inaccuracies due to manual setup further showed the generality of the proposed approach and its robustness.

\section{Conclusion}
\label{sec:conclusion}
We proposed here a new decentralised algorithm to allow robot swarms to build collectively a shared reference system through which every robot can self-localise within the group. The location information can, in turn, be used by the robots to compute their unique coordinates and self-assign specific roles depending on their position. Enabling positional self-awareness 
and location-dependent task allocation can be instrumental to the collective execution of subsequent higher-level operations, such as coordinated motion in formation \citep{pratissoli_soft-bodied_2019} or localised computation \citep{beal2015space}. Our algorithm also enables time coordination of robots' actions through a combination of open- and closed-loop synchronisation mechanisms. Synchronised behaviour is key to achieving effective coordination, both in group-living organisms \citep{Couzin2018synch} as in artificial swarms \citep{Trianni2009,Ghosh2022}.
We showcase our solution in a set of experiments comprising up to 200 real robots that create synchronised dynamic light patterns.

The main distinguishing aspect of our approach compared with previous solutions is the possibility of constructing a shared swarm-level coordinate system with swarms of minimalistic robots, which cannot measure the bearing of other robots. Our algorithms can run on robots only able of basic computation, broadcasting small messages, and making noise-prone estimations of the distance of nearby robots. Instead, most previous work could only achieve decentralised self-localisation through more complex robots equipped with range-and-bearing sensors \citep{beal_organizing_2013,sahin_swarm-bot_2002,guo_swarm_2011,coppola_provable_2019,mathews_mergeable_2017,wang_decentralized_2021,Batra2022,Li2018,Klingner2019}. 
Algorithms for minimalistic platforms have higher portability, due to fewer robot requirements, and can, therefore, be used in a wider range of applications.

We implemented our algorithm on swarms of up to 200 real Kilobot robots, a reference platform for minimalistic collective robotics. 
While our algorithm does not require the ability to detect the bearing of incoming messages, it assumes the robot swarm to be deployed in formation, creating a regular lattice (e.g. a rectangular lattice). The algorithm can thus exploit the regularity of the lattice and its geometric properties to compute the relative location and bearing of each neighbour. Note that, as discussed in Section \ref{sec:SR1c}, the robots do not need to know a priori their formation, rather through subroutine SR1c the robots can deduce online the lattice topology, distinguishing between rectangular and hexagonal lattices. The deployment of large-scale swarms in regular lattices is motivated by the customary practices of storing, charging, and transporting robots organised in such regular formations. This constraint is also in line with previous studies on collective robotics that required the deployment of the robot swarm in a regular lattice \citep{rubenstein_programmable_2014,gauci_programmable_2018,slavkov_morphogenesis_2018,pratissoli_soft-bodied_2019}. Some of these works also ran self-localisation algorithms on Kilobots without using bearing information on messages, however unlike in our approach, they required the use of a small set of robots with specific locations and algorithms to act as `coordinate seeds' \citep{rubenstein_programmable_2014,gauci_programmable_2018}. A further advantage of our algorithm is the absence of predefined roles, rather robots are able to self-determine their role and position at runtime. All robots run the same code and can be freely interchanged.

Our study has a rather practical focus by running experiments with Kilobot swarms in a variety of conditions, lattice formations, and sizes. Robot experiments confirm the possibility of successfully running our algorithm on simple error-prone devices. In order to improve its robustness, we included in our algorithm a set of mechanisms to compensate for individual errors and collectively reach global coordination. However, future work should further increase the algorithm's resiliency by including mechanisms to prevent collective failure when robots break at critical locations of the lattice (e.g., the robot at the coordinates' origin). 
With the proposed algorithm, the swarm would not be impaired by the random deactivation of robots located in the middle of the formation; however, the functioning of robots on the border or the corner of the formation is essential to collective success. Future solutions to compensate for such breakdowns would remove these single points of failure and achieve higher levels of collective fault tolerance.


\section*{Acknowledgements}

This work was supported by OP VVV project International Mobility of Researchers of TBU in Zl\'{i}n project no. CZ.02.2.69/0.0/0.0/16\_027/0008464 and by the resources of A.I.Lab at the Faculty of Applied Informatics, Tomas Bata University in Zl\'{i}n (ailab.fai.utb.cz). A.\,Reina acknowledges financial support from the Belgian F.R.S.-FNRS of which he is a Charg\'{e} de Recherches.

\bibliographystyle{spbasic}  
\bibliography{Biblio.bib}   

\end{document}